%% file: main.tex
\documentclass[10pt,twocolumn,letterpaper]{article}

\usepackage[pagenumbers]{cvpr} 

\input{preamble}

\definecolor{cvprblue}{rgb}{0.21,0.49,0.74}
\usepackage[pagebackref,breaklinks,colorlinks,allcolors=cvprblue]{hyperref}

\usepackage{booktabs}
\usepackage{multirow}
\usepackage[table]{xcolor}

\definecolor{CocoHeader}{HTML}{E8EDF2}
\definecolor{CocoOurs}{HTML}{EAF3FC}
\definecolor{CocoFull}{HTML}{F2F2F2}

\usepackage{booktabs}
\usepackage{array}
\usepackage[table]{xcolor}
\definecolor{CocoHeader}{HTML}{E8EDF2}
\definecolor{CocoOurs}{HTML}{EAF3FC}
\definecolor{CocoFull}{HTML}{F2F2F2}
\newlength{\cocoMetricWidth}
\newcommand{\cocoResult}[2]{\mbox{#1\kern0.08em{\scriptsize$\pm$#2}}}
\newcommand{\cocoBest}[2]{\mbox{\textbf{#1}\kern0.08em{\scriptsize\boldmath$\pm$\textbf{#2}}}}

\definecolor{VocHeader}{HTML}{E8EDF2}
\definecolor{VocOurs}{HTML}{EAF3FC}

\newcommand{\vocBest}[2]
{\mbox{\textbf{#1}\kern0.08em{\scriptsize\boldmath$\pm$\textbf{#2}}}}
\newcommand{\vocResult}[2]{\mbox{#1\kern0.08em{\scriptsize$\pm$#2}}}

\def\confName{CVPR}
\def\confYear{2027}

\title{Mind the Gaps: State-Guided Supervision Allocation for Object Detection Dataset Distillation}

\author{
Jiehui Xie$^{1,*}$ \quad
Zhengxun Jin$^{2,*}$ \quad
Jiahao Zhang$^3$ \quad
Yuxuan Luan$^4$ \\
Dongshen Han$^2$ \quad
Caiyan Qin$^5$ \quad
Sung-Ho Bae$^{2,\dagger}$ \quad
Chaoning Zhang$^{1,\dagger}$ \\[0.6em]
{ $^1$ University of Electronic Science and Technology of China} \\
{ $^2$ Kyung Hee University} \\
{ $^3$ Chengdu University of Information Technology} \\
{ $^4$ Beijing University of Posts and Telecommunications} \\
{ $^5$ Harbin Institute of Technology}
}

\begin{document}
\maketitle

\begingroup
\renewcommand{\thefootnote}{\fnsymbol{footnote}}

\footnotetext[1]{Equal contribution.}

\footnotetext[2]{Corresponding authors:
\texttt{chaoningzhang1990@gmail.com},
\texttt{shbae@khu.ac.kr}.}

\endgroup

\input{sec/0_abstract}    
\input{sec/1_intro}
\input{sec/2_related_work}

\input{sec/3_method}

\input{sec/4_exp}
\input{sec/5_conclusion}

\section*{LLM Usage Statement}

Large language models (LLMs) were used to assist with the writing and refinement of this manuscript. Specifically, LLMs were employed to improve grammatical accuracy, sentence structure, clarity, and overall readability, as well as to enhance the organization and presentation of the text.

{
    \small
    \bibliographystyle{ieeenat_fullname}
    \bibliography{main}
}

\input{sec/X_suppl}

\end{document}

%% file: preamble.tex
\newcommand{\TODO}[1]{\textbf{\color{red}[TODO: #1]}}
\renewcommand{\TODO}[1]{}



%% file: sec/0_abstract.tex
\begin{abstract}
Recent advances in dataset distillation have enabled large-scale training datasets to be compressed into compact synthetic sets. However, dataset distillation for object detection remains challenging, as it requires preserving diverse classification and localization signals under limited image budgets. Existing approaches often emphasize individual-image informativeness without explicitly accounting for supervision accumulated across the distilled dataset, potentially leading to redundant supervision and uneven class--scale coverage. To address this limitation, we propose \textbf{State-Guided Supervision Allocation (SGSA)}, a framework that formulates object detection dataset distillation as a dataset-level progressive supervision allocation problem. SGSA establishes class–scale supervision quotas and maintains a global state that tracks detector predictions accumulated from previously selected synthetic images. The remaining gaps between the target quotas and accumulated supervision guide adaptive object sampling, detector-guided canvas refinement, and marginal-gain-based candidate selection, encouraging complementary supervision across images. On MS COCO, SGSA outperforms a matched OD$^3$ baseline
by 3.73--4.06 mAP points across three image budgets
(0.25\%--1.0\%). On PASCAL VOC, SGSA outperforms the strongest evaluated baseline by 4.53--6.90 mAP$_{50}$ points across three image budgets. These results demonstrate the effectiveness of dataset-level supervision allocation guided by accumulated detector feedback under limited image budgets. 
\end{abstract}

%% file: sec/1_intro.tex
\section{Introduction}
\label{sec:introduction}

\begin{figure}[t]
    \centering
    \includegraphics[width=\columnwidth]{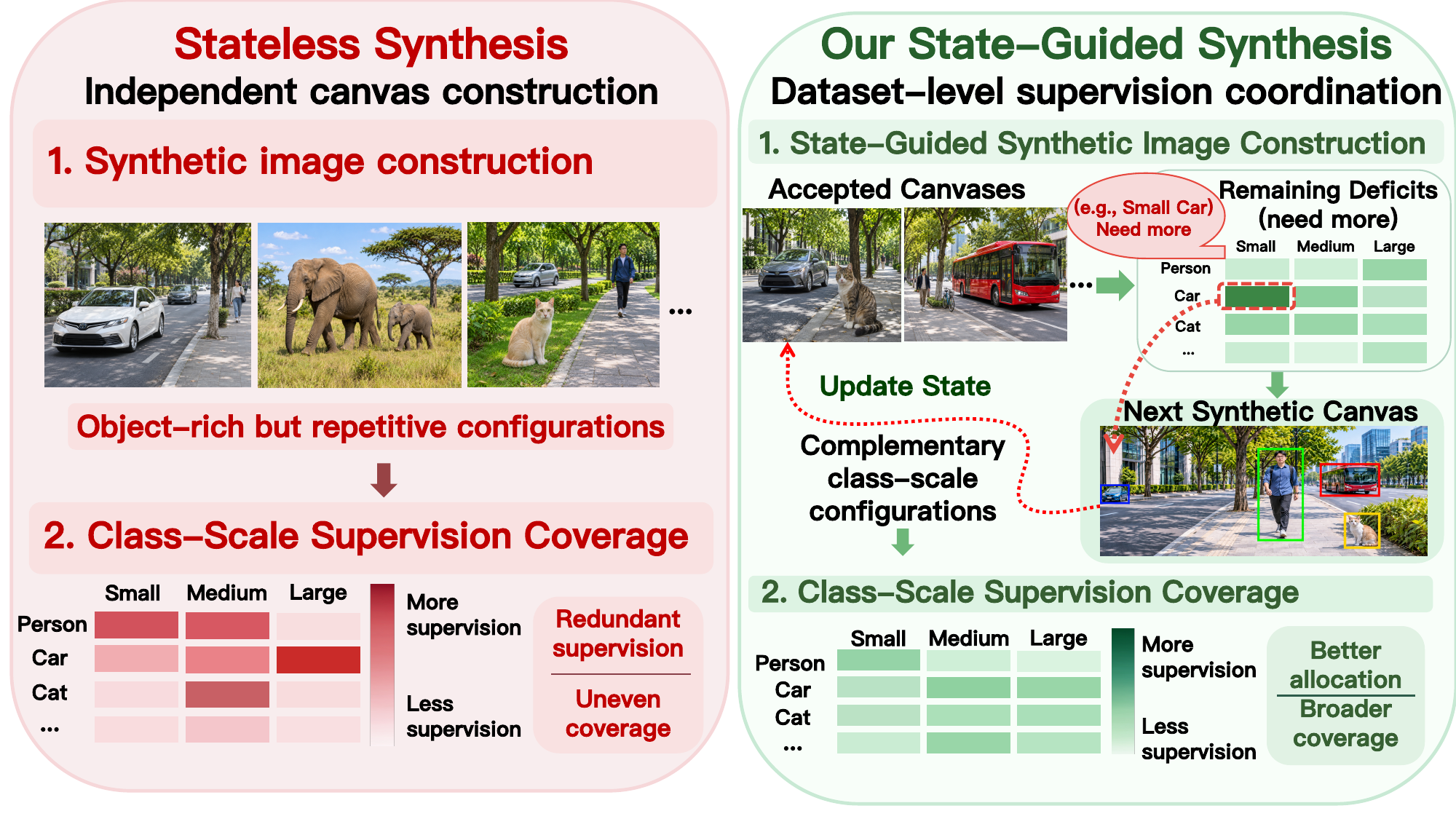}
    \caption{
        \textbf{Motivation of SGSA.}
        \textbf{Left:} Stateless synthesis independently constructs
        images, potentially leading to redundant supervision
        and uneven class--scale coverage.
        \textbf{Right:} Our state-guided synthesis tracks
        global supervision statistics and prioritizes
        under-covered class--scale configurations,
        encouraging complementary supervision
        under a limited image budget.
    }
    \label{fig:motivation}
\end{figure}

Object detection has achieved remarkable progress with the development of deep neural networks and large-scale annotated datasets \cite{ren2015faster,lin2017feature,carion2020end,zhang2023dino}. However, achieving strong detection performance typically
requires large amounts of annotated training data and
substantial computational resources, particularly when dealing with diverse object categories, scales, and spatial configurations \cite{singh2018sniper}. Dataset distillation (DD) \cite{wang2018dataset,yu2024review} offers a promising solution by compressing large training datasets into compact synthetic sets while preserving their effectiveness for downstream training. For image classification, dataset distillation has shown
promising results through gradient matching
\cite{zhao2020dataset,zhao2021dataset,kim2022idc}, trajectory matching
\cite{cazenavette2022dataset,du2023ftd,guo2024datm}, and distribution matching
\cite{zhao2023dataset,zhao2023improved}.
Extending dataset distillation to object detection requires
preserving both semantic and localization information
across multiple object instances.

Recent detection dataset distillation methods explore model inversion \cite{qi2024fetch}, real-content composition \cite{al2026od,hu2026focusdc}, and task-guided diffusion \cite{qi2025unidd,chen2026mgpo} to construct informative synthetic training images. Beyond image construction, RCE \cite{tran2026rce} addresses category imbalance through rarity-guided caption expansion, encouraging the generation of underrepresented object categories. While these approaches improve image informativeness or content coverage, they do not necessarily account for supervision already accumulated across the distilled dataset. In particular, a synthetic image containing numerous detectable objects may still contribute supervision that is largely redundant with previously selected images. As illustrated in Fig.~\ref{fig:motivation}, constructing images without considering accumulated supervision may repeatedly favor certain categories and scales, leaving other configurations insufficiently represented under a limited image budget.

This motivates us to view detection dataset distillation as a dataset-level supervision allocation problem. Rather than evaluating synthetic images solely by their individual informativeness, effective dataset construction should also consider the supervision already represented by previously selected images. This perspective is related to coverage-aware coreset selection \cite{sener2017active,lee2024coreset,pooladzandi2022adacore}, diversity-aware sampling \cite{yang2024plug,wu2022entropy,ash2019deep}, and class-balanced learning \cite{cui2019class,tan2021eqlv2}. However, sequential synthetic image construction introduces an additional consideration: each accepted image changes the supervision requirements for subsequent images. This raises a central question: \textit{How can accumulated detector feedback guide the construction of successive synthetic images toward complementary supervision under a limited image budget?}

To this end, we propose \textbf{SGSA}, a state-guided framework for object detection dataset distillation that coordinates synthetic image construction through global supervision feedback. Specifically, we establish class–scale supervision quotas and maintain a global state that accumulates class–scale detection counts from a pre-trained detector, termed the \textit{Observer}, over previously selected synthetic images. The remaining deficits between the target quotas and accumulated supervision guide adaptive object sampling and canvas construction. Observer predictions guide the filtering and refinement of composed canvases, while marginal-gain-based selection favors candidates that best address the remaining deficits. After each selection, the accepted candidate's detection counts update the global state, adjusting the sampling priorities for subsequent images. This progressive process encourages complementary supervision across the distilled dataset. The resulting images are subsequently re-annotated and used for detector training under an established feature distillation protocol.

Experiments on MS COCO and PASCAL VOC demonstrate the effectiveness of SGSA under limited image budgets. On MS COCO, SGSA outperforms a matched OD$^3$ baseline
by 3.73--4.06 mAP points across three image budgets
(0.25\%--1.0\%). On PASCAL VOC, SGSA improves mAP$_{50}$ by 4.53--6.90 points over the strongest evaluated baseline across three image budgets. Ablation studies identify deficit-driven sampling as the main contributor to performance gains and examine the effects of state feedback and supervision granularity. Further analysis shows that SGSA improves class--scale quota utilization compared with natural sampling.

Our main contributions are summarized as follows:
\begin{itemize}
    \item To the best of our knowledge, we are the first to formulate detection dataset distillation as a \textbf{dataset-level progressive supervision allocation problem}, accounting for supervision accumulated across synthetic images.

    \item We propose \textbf{SGSA}, a state-guided framework that coordinates synthetic image construction through accumulated detector feedback and remaining class--scale supervision deficits.

    \item Experiments on MS COCO and PASCAL VOC demonstrate \textbf{consistent performance gains} across image budgets, with ablations examining the contributions of supervision allocation and state feedback.
\end{itemize}

%% file: sec/2_related_work.tex
\section{Related Work}
\label{sec:rw}

\begin{figure*}[t]
    \centering
    \includegraphics[width=\textwidth]{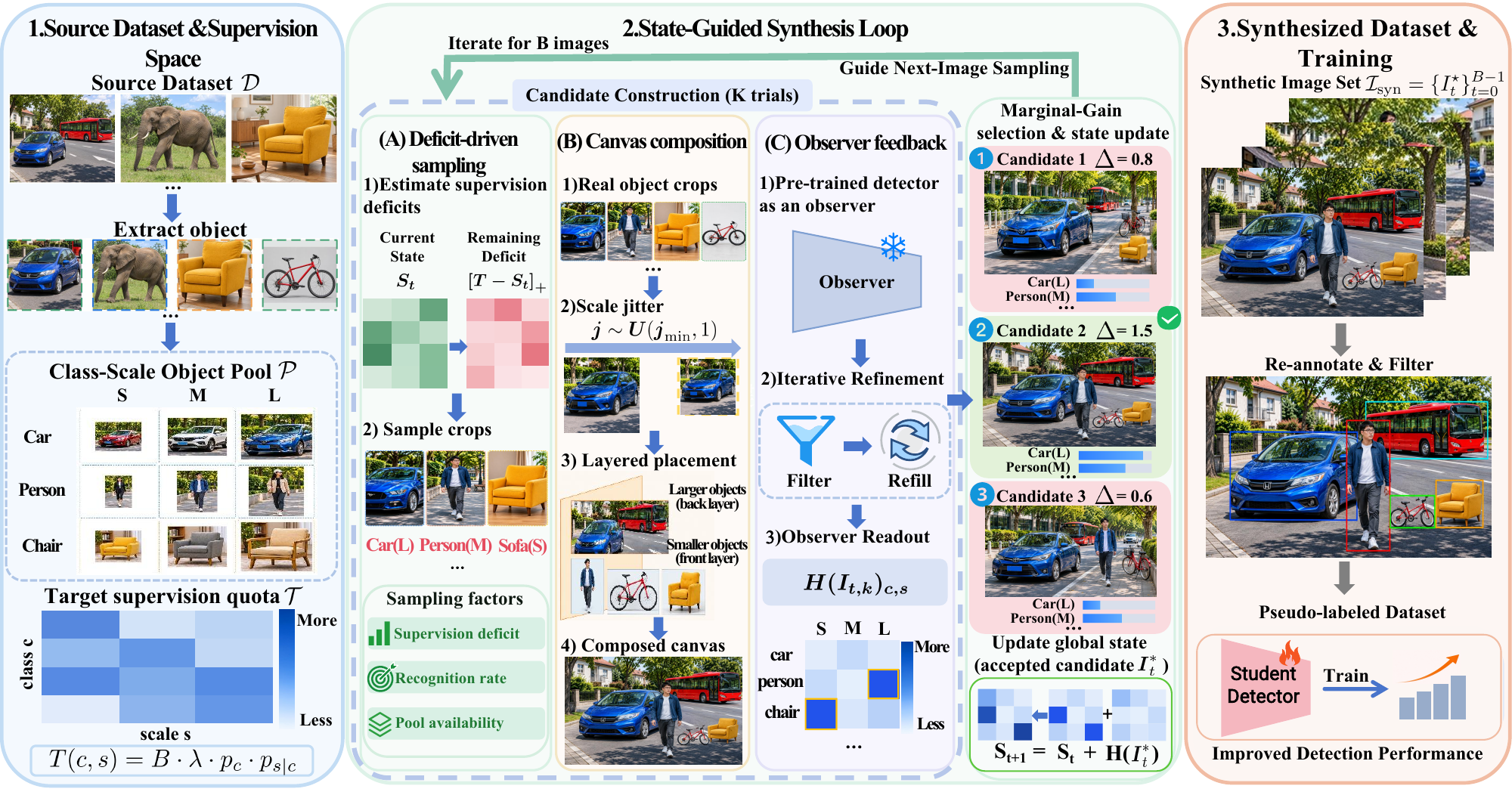}
    \caption{
        \textbf{Overview of SGSA.}
        Our method consists of three main stages:
        (1) constructing a class--scale indexed object pool
        and defining target supervision quotas;
        (2) iteratively synthesizing detection images through
        deficit-driven object sampling, real-object composition,
        observer feedback, and marginal-gain-based candidate
        selection with global state updates;
        and (3) re-annotating the synthesized dataset and
        training a student detector with pseudo-label supervision
        and feature distillation.
        SGSA coordinates supervision across images without generative models. The frozen Observer also serves as the teacher for pseudo-labeling and feature distillation.
    }
    \label{fig:framework}
\end{figure*}

\subsection{Dataset Distillation for Object Detection}
Dataset distillation compresses large datasets into compact synthetic sets while preserving training effectiveness \cite{wang2018dataset}. For object detection, this requires retaining semantic, localization, and scale information across multiple objects. DCOD \cite{qi2024fetch} synthesizes detection-oriented images through model inversion with foreground--background decoupling. OD$^3$ \cite{al2026od} adopts optimization-free object composition and detector-guided screening, while FocusDC \cite{hu2026focusdc} combines salient patches with downsampled image views to preserve informative content and scene context. Diffusion-based approaches offer another direction: UniDD \cite{qi2025unidd} employs task-specific proxy models to guide generation, RCE \cite{tran2026rce} expands prompts toward underrepresented categories, and MGPO \cite{chen2026mgpo} incorporates detector-derived rewards and latent geometric regularization. These methods advance detection dataset distillation through image construction, task alignment, and content coverage. Beyond the quality of individual synthetic images, effective distillation under limited budgets also requires considering how class--scale supervision is distributed across the dataset. Our work addresses this complementary problem by coordinating supervision across successive synthetic images to promote more balanced and complementary training signals.

\subsection{Coverage-Aware Data Selection and Sampling}
Data selection under limited budgets emphasizes representativeness \cite{sener2017active}, gradient preservation \cite{mirzasoleiman2020coresets,killamsetty2021grad}, and uncertainty--diversity trade-offs \cite{ash2019deep}. In classification-oriented dataset distillation, DREAM \cite{liu2023dream} selects representative real samples for matching, while SelMatch \cite{lee2024selmatch} combines selection-based initialization with partial trajectory matching. For object detection, CALD \cite{yu2022consistency} and PPAL \cite{yang2024plug} employ active learning to select informative images, whereas CSOD \cite{lee2024coreset} constructs representative subsets using classwise object features. Related studies on class-balanced learning \cite{cui2019class,tan2021eqlv2}, long-tailed dataset distillation \cite{zhao2025ltdd,cui2026rethinking}, and scale-aware detection \cite{singh2018snip,singh2018sniper} further highlight the importance of category and scale distributions. While these studies emphasize informative sample selection and balanced data representations, sequential synthetic dataset construction additionally requires accounting for supervision already provided by previously selected images. SGSA addresses this need by tracking accumulated class--scale supervision in a global state and using remaining deficits relative to target quotas to guide subsequent object sampling and candidate selection.

%% file: sec/3_method.tex
\section{Method}
\label{sec:method}

We present \textbf{SGSA}, a state-guided framework for
object detection dataset distillation.
As illustrated in Fig.~\ref{fig:framework}, our framework
coordinates object sampling and candidate selection through
a global category--scale supervision state, guiding successive
images to address remaining supervision deficits.
The resulting images are pseudo-labeled and used to train
a student detector.

\subsection{Problem formulation.}
Given a labeled detection dataset
$\mathcal{D}_{\mathrm{real}}=\{(x_i,y_i)\}_{i=1}^{N}$
and an image budget $B\ll N$, we aim to construct a compact
synthetic dataset
$\mathcal{D}_{\mathrm{syn}}=\{(I_i,\hat y_i)\}_{i=0}^{B-1}$,
where $\hat y_i$ denotes teacher-generated detection annotations.
The goal is to enable a student detector trained on
$\mathcal{D}_{\mathrm{syn}}$ to generalize to real images.
Each synthetic image can contain multiple objects across categories
and scales, so a fixed image budget does not uniquely determine
the amount or composition of supervision. We therefore study class--scale supervision allocation
to capture semantic diversity and scale-dependent
detection challenges under this budget.

\subsection{Category--Scale Supervision Targets}
\label{sec:supervision_targets}

\noindent\textbf{Object pools.}
We organize source objects into category--scale pools
$\{\mathcal{P}_{c,s}\}$, where $c\in\{1,\ldots,C\}$ denotes an object
category and $s\in\{0,1,2\}$ denotes a scale bin. To account for
differences in source-image resolution, we first map object dimensions
to a reference canvas of size $W\times H$. For an object with
bounding-box dimensions $(w,h)$ in a source image of size
$W_{\mathrm{src}}\times H_{\mathrm{src}}$, the mapped dimensions are
\begin{equation}
    (\bar w,\bar h)=(fw,fh),
    \qquad
    f=\sqrt{\frac{WH}{W_{\mathrm{src}}H_{\mathrm{src}}}}.
    \label{eq:reference_scaling}
\end{equation}

This transformation preserves the object-to-image
area ratio before subsequent augmentation.
We assign each source object to a scale bin
$s=\operatorname{bin}(a)$ according to its mapped
bounding-box area $a=\bar w\bar h$:
\begin{equation}
    \operatorname{bin}(a)
    =
    \begin{cases}
        0, & a<\tau_1^2,\\
        1, & \tau_1^2\leq a<\tau_2^2,\\
        2, & a\geq\tau_2^2,
    \end{cases}
    \label{eq:scale_bins}
\end{equation}
where $0<\tau_1<\tau_2$ are the scale
thresholds defined on the reference canvas.

Each source object $o_i$ with category $c_i$
and mapped bounding-box area $a_i$ is assigned
to the pool
$\mathcal{P}_{c_i,\operatorname{bin}(a_i)}$.
We retain surrounding image context when
extracting object patches.

\noindent\textbf{Target allocation.}
Let $n_{c,s}=|\mathcal{P}_{c,s}|$ denote the number of
source objects in category--scale pool $\mathcal{P}_{c,s}$,
and $N_c=\sum_s n_{c,s}$ denote the total number of
source objects in category $c$.
We define a softened category distribution
and a category-conditional scale distribution as
\begin{equation}
    p_c
    =
    \frac{N_c^\alpha}{\sum_{c'}N_{c'}^\alpha},
    \qquad
    p_{s\mid c}
    =
    \frac{n_{c,s}}{N_c},
    \label{eq:target_distribution}
\end{equation}
where $0<\alpha\leq1$ controls the influence
of category frequency.

The target supervision for each category--scale cell is
\begin{equation}
    T(c,s)=B\lambda\,p_c\,p_{s\mid c},
    \label{eq:target}
\end{equation}
where $\lambda$ specifies the target amount of supervision
per synthetic image, measured by predictions from a
frozen pre-trained detector, termed the \emph{Observer}. Consequently,
$\sum_{c,s}T(c,s)=B\lambda$. This allocation reduces the dominance of frequent categories while preserving the observed scale proportions within each category. The resulting soft targets guide subsequent object sampling and candidate selection.

\subsection{Global State and Deficit-Driven Sampling}
\label{sec:state_sampling}

\noindent\textbf{Supervision state.}
Let $S_t\in\mathbb{R}_{\geq0}^{C\times3}$ denote the
accumulated class--scale supervision counts after
accepting $t$ synthetic images, with $S_0=\mathbf{0}$.
Each entry $S_t(c,s)$ counts high-confidence Observer
predictions for category $c$ and scale bin $s$.
For a canvas $I$, the Observer produces predictions
\begin{equation}
    \mathcal{O}(I)
    =
    \{(\hat b_j,\hat c_j,q_j)\}_{j=1}^{m(I)},
    \label{eq:observer_predictions}
\end{equation}
where $m(I)$ is the number of predictions,
and $\hat b_j$, $\hat c_j$, and $q_j$
denote the predicted box, category, and confidence
of the $j$-th prediction, respectively. We summarize its predictions by
\begin{equation}
    H(I)_{c,s}
    =
    \sum_{j=1}^{m(I)}
    \mathbf{1}\!\left[
        q_j \geq \gamma_{\mathrm{gain}},\;
        \hat c_j = c,\;
        \operatorname{bin}(|\hat b_j|) = s
    \right]
    \label{eq:histogram}
\end{equation}
where $\mathbf{1}[\cdot]$ is the indicator function,
and $|\hat b_j|$ denotes the predicted bounding-box area.

This histogram counts high-confidence Observer predictions
by category and scale without requiring correspondence
to individual pasted objects.

\noindent\textbf{Empirical recognition rate.}
Pasting an object does not guarantee that the Observer will recognize
it. We therefore maintain a smoothed empirical recognition rate for
each pool:
\begin{equation}
    \rho(c,s)
    =
    \frac{
        \mathrm{succ}(c,s)+\kappa p_0
    }{
        \mathrm{trials}(c,s)+\kappa
    },
    \label{eq:recognition_rate}
\end{equation}
where $p_0$ is the prior recognition rate and $\kappa$ controls
smoothing. The counters record evaluated object attempts and successful
recognitions, respectively. An evaluated object is considered
successfully recognized if there exists an Observer prediction with
the same category, confidence at least $\gamma_{\mathrm{rec}}$, and IoU of at least
$0.5$ with its transformed source box.

\noindent\textbf{Deficit-driven planning.}
During a planning call, let $P^{(\ell)}(c,s)$ denote the number of
objects selected from cell $(c,s)$ before sampling step $\ell$.
The planner initializes $P^{(0)}=\mathbf{0}$ at the beginning of each
call, including refill calls. We estimate the remaining supervision
deficit as
\begin{equation}
    d_t^{(\ell)}(c,s)
    =
    \bigl[
        T(c,s)-S_t(c,s)
        -P^{(\ell)}(c,s)\rho(c,s)
    \bigr]_+
    \label{eq:deficit}
\end{equation}
where $[x]_+=\max(x,0)$. The last term accounts for the estimated
contribution of objects already selected within the current call.

We convert these deficits into sampling weights:
\begin{equation}
    w_t^{(\ell)}(c,s)
    =
    \frac{
        d_t^{(\ell)}(c,s)
    }{
        \max(T(c,s),\epsilon)
    }
    \cdot
    \rho(c,s)\,M^{(\ell)}(c,s)
    \label{eq:sampling_weights}
\end{equation}
where $\epsilon>0$ ensures numerical stability. The binary mask
$M^{(\ell)}$ encodes pool availability and the planning limit on
large-scale objects. When the weights have positive total mass,
the next category--scale cell is sampled from
\begin{equation}
    \pi_t^{(\ell)}(c,s)
    =
    \frac{
        w_t^{(\ell)}(c,s)
    }{
        \sum_{c',s'}w_t^{(\ell)}(c',s')
    }.
    \label{eq:sampling_probability}
\end{equation}
where the denominator sums the sampling weights
over all category--scale cells. An object is then drawn from the selected pool, and the corresponding
entry of $P^{(\ell)}$ is incremented.

Since sampled objects may not yield the expected
detection supervision due to occlusion or recognition
failures, fixed sampling may leave class--scale gaps.
SGSA instead adjusts sampling priorities based on
accumulated Observer counts and empirical recognition
rates, progressively addressing the remaining deficits
across successive canvases.

\subsection{Observer-Guided Canvas Construction}
\label{sec:canvas_construction}

\noindent\textbf{Scale augmentation and layout.}
We construct candidate canvases from the selected real-image patches.
Each context patch undergoes independent isotropic scale jitter,
\begin{equation}
    \begin{aligned}
        j_i
        &\sim \mathcal{U}(j_{\min},1),\\
        (w_i',h_i')
        &=
        \left(
            j_iw_i^{\mathrm{patch}},
            j_ih_i^{\mathrm{patch}}
        \right),
    \end{aligned}
    \label{eq:scale_jitter}
\end{equation}
where $j_i$ is the scale factor for patch $i$,
$j_{\min}$ is the minimum scale factor,
and $(w_i^{\mathrm{patch}},h_i^{\mathrm{patch}})$
and $(w_i',h_i')$ denote the patch dimensions
before and after scaling, respectively.
The same transformation is applied to its
associated object box.

We use a two-layer layout, assigning larger patches to the back layer
and smaller patches to the front layer according to their jittered patch
areas. Within each layer, patches are considered in descending area
order. For a candidate patch region $R_i$, let $U_{\mathrm{layer}}$
denote the region already occupied in that layer. We measure occupancy
by
\begin{equation}
    \operatorname{Occ}(R_i)
    =
    \frac{
        |R_i\cap U_{\mathrm{layer}}|
    }{
        |R_i|
    }.
    \label{eq:occupancy}
\end{equation}

A placement is allowed only when this ratio satisfies the corresponding
layer-specific threshold. Front-layer placements are additionally
constrained to limit occlusion of back-layer objects.

\noindent\textbf{Filtering and refill.}
After rendering a canvas, we evaluate it with the Observer. A pasted
object is retained if its transformed source box overlaps at least one
predicted box with IoU greater than $0.25$. This retention criterion
does not require category agreement or impose an additional confidence
threshold.

Objects that fail the retention criterion are removed. We then invoke
the deficit-driven planner again to select replacement objects,
allowing refill to depend on the current dataset state. We perform $R$ refill rounds, followed by
a final rendering and Observer evaluation. The resulting final predictions are used for
candidate scoring and supervision accounting.

\subsection{Marginal-Gain Selection and State Feedback}
\label{sec:candidate_selection}

\begin{table*}[t]
  \centering
  \small
  \setlength{\tabcolsep}{2.4pt}
  \newcommand{\cocoBudgetSep}{{\color{black!30}\vrule width 0.5pt\relax}}
  \setlength{\cocoMetricWidth}{%
    \dimexpr(\textwidth-0.12\textwidth-20\tabcolsep-1.5pt)/9\relax}
  \renewcommand{\arraystretch}{1.18}
  \caption{Object detection performance on MS COCO under three image budgets.
All metrics are reported in \%. We use the scores reported in
OD$^3$~\cite{al2026od} for Random, Uniform, K-Center,
Herding, and DCOD, and those in UniDD~\cite{qi2025unidd}
for UniDD (AP$_{75}$ not reported).
Results for our reproductions of OD$^3$ and OD$^3$+gate,
and for SGSA, are reported as mean $\pm$ sample standard
deviation across three runs.}
  \label{tab:coco_main}
  \begin{tabular}{p{0.12\textwidth}!{\cocoBudgetSep}
    *{3}{>{\centering\arraybackslash}p{\cocoMetricWidth}}!{\cocoBudgetSep}
    *{3}{>{\centering\arraybackslash}p{\cocoMetricWidth}}!{\cocoBudgetSep}
    *{3}{>{\centering\arraybackslash}p{\cocoMetricWidth}}}
    \toprule
    \rowcolor{CocoHeader}
    \textbf{Method}
    & \multicolumn{3}{c!{\cocoBudgetSep}}{\textbf{0.25\% budget}}
    & \multicolumn{3}{c!{\cocoBudgetSep}}{\textbf{0.5\% budget}}
    & \multicolumn{3}{c}{\textbf{1.0\% budget}} \\
    \cmidrule(lr){2-4}\cmidrule(lr){5-7}\cmidrule(lr){8-10}
    & mAP & AP$_{50}$ & AP$_{75}$
    & mAP & AP$_{50}$ & AP$_{75}$
    & mAP & AP$_{50}$ & AP$_{75}$ \\
    \midrule
    Random
    & 3.50 & 9.70 & 1.60
    & 5.50 & 14.20 & 2.90
    & 8.30 & 19.70 & 5.30 \\
    Uniform
    & 3.60 & 9.80 & 1.60
    & 5.60 & 14.30 & 2.90
    & 8.40 & 19.70 & 5.40 \\
    K-Center
    & 1.70 & 6.30 & 0.40
    & 2.80 & 8.90 & 0.70
    & 4.00 & 12.90 & 1.20 \\
    Herding
    & 1.70 & 5.80 & 0.50
    & 2.60 & 8.80 & 0.80
    & 4.10 & 12.50 & 1.30 \\
    DCOD
    & 7.20 & 17.20 & 4.80
    & 10.00 & 21.50 & 8.00
    & 12.10 & 24.70 & 10.40 \\
    UniDD
    & \cocoResult{4.50}{0.30} & \cocoResult{10.30}{0.40} & --
    & \cocoResult{7.10}{0.40} & \cocoResult{16.90}{0.30} & --
    & \cocoResult{10.80}{0.40} & \cocoResult{22.50}{0.50} & -- \\
    OD$^3$
    & \cocoResult{12.00}{0.20} & \cocoResult{23.53}{0.35} & \cocoResult{10.93}{0.15}
    & \cocoResult{16.77}{0.15} & \cocoResult{31.70}{0.17} & \cocoResult{15.90}{0.26}
    & \cocoResult{20.67}{0.06} & \cocoResult{37.67}{0.06} & \cocoResult{20.53}{0.25} \\
    OD$^3$+gate
    & \cocoResult{12.33}{0.23} & \cocoResult{23.93}{0.42} & \cocoResult{11.47}{0.15}
    & \cocoResult{17.27}{0.42} & \cocoResult{32.07}{0.58} & \cocoResult{16.77}{0.55}
    & \cocoResult{21.97}{0.15} & \cocoResult{38.83}{0.38} & \cocoResult{22.33}{0.23} \\
    \midrule
    \rowcolor{CocoOurs}
    \textbf{SGSA (Ours)}
    & \cocoBest{15.80}{0.30} & \cocoBest{30.10}{0.56} & \cocoBest{15.07}{0.21}
    & \cocoBest{20.50}{0.35} & \cocoBest{37.30}{0.61} & \cocoBest{20.40}{0.17}
    & \cocoBest{24.73}{0.12} & \cocoBest{42.97}{0.06} & \cocoBest{25.60}{0.17} \\
    \bottomrule
  \end{tabular}
\end{table*}

\begin{table*}[t]
  \centering
  \captionsetup{position=top,skip=5pt,width=\textwidth}
  \caption{Object detection performance on PASCAL VOC at three
  image budgets, measured in mAP$_{50}$ (\%).
  Baseline results are from OD$^3$~\cite{al2026od}
  and UniDD~\cite{qi2025unidd}.
  We report mean $\pm$ uncertainty estimates where available.
  For OD$^3$, approximate sample standard deviations are
  derived from the standard errors reported over four runs.
  For SGSA, mean and sample standard deviation are computed
  over three runs.}
  \label{tab:voc_main}
  \small
  \setlength{\tabcolsep}{4.5pt}
  \renewcommand{\arraystretch}{1.18}
  \begin{tabular}{c*{7}{c}>{\columncolor{VocOurs}}c}
    \toprule
    \rowcolor{VocHeader}
    \textbf{Budget}
    & \textbf{Random}
    & \textbf{Herding}
    & \textbf{K-Center}
    & \textbf{Uniform}
    & \textbf{DCOD}
    & \textbf{UniDD}
    & \textbf{OD$^3$}
    & \textbf{SGSA (Ours)} \\
    \midrule
    0.5\%
    & 15.80 & 12.60 & 14.50 & 15.80 & 37.90
    & \vocResult{22.30}{0.60}
    & \vocResult{38.50}{0.20}
    & \vocBest{43.03}{0.65} \\
    1.0\%
    & 25.50 & 19.30 & 21.90 & 25.70 & 46.40
    & \vocResult{38.90}{0.70}
    & \vocResult{51.10}{0.40}
    & \vocBest{57.80}{0.75} \\
    2.0\%
    & 40.50 & 28.10 & 31.30 & 40.60 & 50.70
    & \vocResult{48.50}{0.60}
    & \vocResult{58.70}{0.20}
    & \vocBest{65.60}{0.36} \\
    \bottomrule
  \end{tabular}
\end{table*}

For each dataset position $t$, we construct a candidate set
$\mathcal{C}_t=\{I_{t,1},\ldots,I_{t,K}\}$. Candidates are generated
sequentially using shared construction resources, including the object
pools and random-number stream.

To evaluate a candidate's contribution to the dataset, we define the
quota-capped supervision utility
\begin{equation}
    Q(S)
    =
    \sum_{c,s}
    \min\bigl(S(c,s),T(c,s)\bigr).
    \label{eq:utility}
\end{equation}

Its marginal gain at state $S_t$ is
\begin{equation}
    \begin{aligned}
        \Delta Q(I\mid S_t)
        &=
        Q\bigl(S_t+H(I)\bigr)-Q(S_t)
        \\
        &=
        \sum_{c,s}
        \min\!\left(
            H(I)_{c,s},
            [T(c,s)-S_t(c,s)]_+
        \right).
    \end{aligned}
    \label{eq:marginal_gain}
\end{equation}

This score credits predictions up to the remaining deficit in each
category--scale cell. Predictions beyond an already satisfied target
receive no additional reward. Candidate evaluation therefore accounts
for the supervision accumulated in previous images.

We accept the candidate with the largest marginal gain and update the
global state:
\begin{equation}
        I_t^\star=
        \underset{I\in\mathcal{C}_t}{\arg\max}\;
        \Delta Q(I\mid S_t),
        S_{t+1}
        =
        S_t+H(I_t^\star).
    \label{eq:state_update}
\end{equation}

Only the accepted candidate contributes to $S_t$. Although the utility
caps contributions at the target, the state update records the complete
histogram, including predictions exceeding their corresponding quotas.
Construction continues until $B$ images have been accepted, including
when the candidate gains are zero. Thus, the targets guide supervision
allocation without guaranteeing that every quota is fulfilled.

This feedback allows subsequent sampling to adapt to observed class–scale supervision rather than relying solely on a fixed allocation.

\subsection{Pseudo-Labeling and Student Training}
\label{sec:student_training}

Once construction is complete, we obtain
the synthesized image set
\[
    \mathcal{I}_{\mathrm{syn}}
    =
    \{I_t^\star\}_{t=0}^{B-1}.
\]

We relabel these images using the Observer
with test-time augmentation.
For each image, we retain the highest-ranked
predictions and apply confidence filtering
to obtain pseudo annotations
$\hat y_t=\{(\hat b_j,\hat c_j)\}$.
The resulting pseudo-labeled dataset is $\mathcal{D}_{\mathrm{syn}}
=\{(I_t^\star,\hat y_t)\}_{t=0}^{B-1}$. These annotations are generated separately
from the construction-time histograms in
Eq.~\eqref{eq:histogram}.
The student uses the resulting boxes and
hard category labels; prediction confidences
are not used as soft class targets.

We then train the student using the standard Faster R-CNN
detection loss,
\begin{equation}
    \mathcal{L}_{\mathrm{det}}
    =
    \mathcal{L}_{\mathrm{RPN}}
    +
    \mathcal{L}_{\mathrm{RoI}},
    \label{eq:detection_loss}
\end{equation}
where the RPN and RoI losses include the standard
classification and box-regression terms. To provide additional feature-level supervision,
we adopt PKD feature distillation
\cite{NEURIPS2022_631ad9ae},
which uses normalized feature matching between
the student and teacher FPN features.
The overall training objective is
\begin{equation}
    \mathcal{L}_{\mathrm{train}}
    =
    \mathcal{L}_{\mathrm{det}}
    +
    \sum_{l=0}^{L-1}
    \eta_l
    \mathcal{L}_{\mathrm{PKD}}
    \left(
        F_{\mathrm{stu}}^{(l)},
        F_{\mathrm{tea}}^{(l)}
    \right),
    \label{eq:training_loss}
\end{equation}
where $L$ is the number of FPN levels used for
distillation, $F_{\mathrm{stu}}^{(l)}$ and
$F_{\mathrm{tea}}^{(l)}$ denote the student
and teacher features at level $l$,
respectively, and $\eta_l$ controls the
corresponding distillation weight.

%% file: sec/4_exp.tex
\section{Experiments}

\subsection{Experimental Setup}
\label{sec:experimental_setup}

\paragraph{Datasets and evaluation.}
We evaluate SGSA on MS COCO \cite{lin2014microsoft} and PASCAL VOC \cite{everingham2010pascal}. The image budget is defined as the ratio of synthetic images to the size of the corresponding original training set, expressed as a percentage.
For COCO, we construct distilled datasets from
train2017 and evaluate on val2017,
using image budgets of 0.25\%, 0.5\%, and 1\%.
For VOC, we use the combined VOC2007 and VOC2012
trainval sets for construction and
VOC2007 test for evaluation.
The corresponding budgets are 0.5\%, 1\%, and 2\%.
We report COCO mAP averaged over IoU thresholds
from 0.50 to 0.95, together with AP$_{50}$ and
AP$_{75}$.
For VOC, we report mAP$_{50}$ at an IoU threshold of 0.50, computed using COCO-style evaluation on converted VOC annotations.

\paragraph{Baselines.}
We compare SGSA with four approaches: real-image subset selection (Random~\cite{rebuffi2017icarl}, Uniform~\cite{lee2024coreset}, K-Center~\cite{sener2017active}, Herding~\cite{3023549.3023562}), model inversion (DCOD~\cite{qi2024fetch}), diffusion-based synthesis (UniDD~\cite{qi2025unidd}), and object composition (OD$^3$~\cite{al2026od}), at matched image budgets on MS COCO and PASCAL VOC. On MS COCO, we also evaluate OD$^3$+gate using SGSA's pseudo-label confidence threshold (0.3) for a controlled comparison.

\paragraph{Implementation details.}
We use Faster R-CNN R50-FPN students with
ImageNet-pretrained backbones.
Pretrained Faster R-CNN detectors with R101-FPN
on COCO and R50-FPN on VOC serve as the Observer,
pseudo-labeler, and PKD teacher.
For dataset construction, we set
$\alpha=0.5$, $\lambda=50$, $K=2$, and $R=2$.
Students are trained for $96$ epochs using SGD
with a total batch size of four.
PKD feature distillation is applied to
$L=4$ FPN levels with $\eta_l=6$,
unless otherwise specified.
Results are reported as mean $\pm$ sample
standard deviation over three runs with paired
synthesis and training seeds. Additional implementation details are provided
in the supplementary material.

\subsection{Main Results}
\label{sec:main_results}

\paragraph{Results on MS COCO.}
Table~\ref{tab:coco_main} presents detection results across three image budgets. SGSA consistently achieves the best performance across all budgets and evaluation metrics. Specifically, it obtains 15.80, 20.50, and 24.73 mAP at the 0.25\%, 0.5\%, and 1.0\% budgets, outperforming the strongest matched baseline, OD$^3$+gate, by 3.47, 3.23, and 2.76 mAP points, respectively. Notably, the largest improvement occurs at the smallest budget, corresponding to a 28.1\% relative mAP gain. Consistent improvements in $\mathrm{AP}_{50}$ and $\mathrm{AP}_{75}$ further demonstrate the effectiveness of SGSA under different IoU thresholds. These results highlight the strong advantage of SGSA in extremely data-constrained settings.
\paragraph{Results on PASCAL VOC.}
Table~\ref{tab:voc_main} presents detection results on PASCAL VOC, where SGSA achieves the best $\mathrm{mAP}_{50}$ across all three image budgets. Specifically, SGSA obtains $\mathrm{mAP}_{50}$ scores of 43.03, 57.80, and 65.60 at the 0.5\%, 1.0\%, and 2.0\% budgets, outperforming the strongest baseline, OD$^3$, by 4.53, 6.70, and 6.90 points, respectively. These consistent gains complement the COCO results and further demonstrate the effectiveness of SGSA across different datasets and image budgets.
\begin{figure*}[t]
    \centering
    \includegraphics[width=\textwidth]{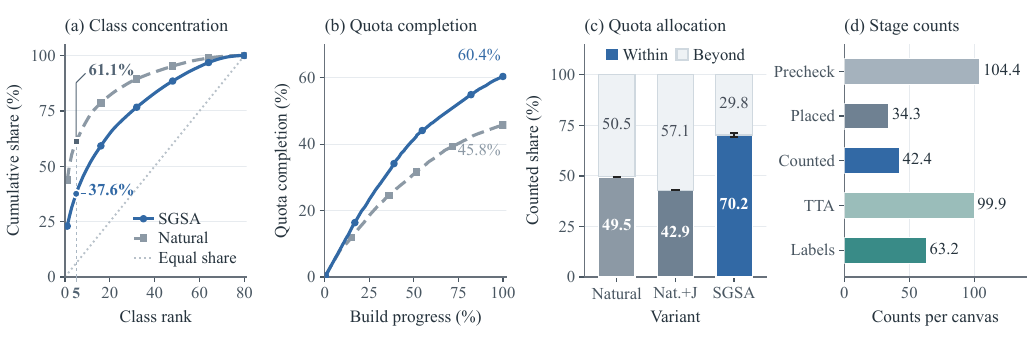}
    \caption{
    \textbf{Supervision distribution and quota completion
    on COCO at a 0.25\% image budget.}
    (a) Cumulative category supervision share, with categories
    sorted separately for each method; the dotted line
    indicates uniform allocation.
    (b) Capped quota completion during construction.
    Panels (a,b) show seed-0 runs.
    (c) Within-quota share,
    $\sum_{c,s}\min(S(c,s),T(c,s))/\sum_{c,s}S(c,s)$,
    and complementary beyond-quota share.
    Bars show means over three seeds, with error bars
    indicating sample standard deviations.
    Shares reflect quota accounting rather than label accuracy.
    Natural uses no jitter, while Nat.+J matches SGSA's
    scale jitter $[0.5,1]$.
    Both natural variants use count-based selection,
    whereas SGSA uses deficit-driven sampling and selection.
    (d) Mean per-canvas counts for SGSA (seed 0):
    objects before verification (Precheck),
    final placed objects (Placed),
    single-pass Observer predictions (Counted),
    top-100 TTA predictions (TTA),
    and pseudo-labels filtered at a 0.3 score threshold (Labels).
}
    \label{fig:mechanism}
\end{figure*}

\subsection{Ablation Studies}
\label{sec:ablation}

We conduct ablations on COCO at a 0.25\% image budget using a Faster R-CNN R50 student trained for 96 epochs. Unless otherwise specified, construction and training settings remain unchanged. We report mean mAP and sample standard deviation over three runs with paired synthesis and training seeds.

\paragraph{Deficit-driven sampling and candidate selection.}
We independently vary object sampling and candidate selection
in a $2\times2$ design, retaining the same layout and scale jitter
in all four configurations.
As shown in Table~\ref{tab:ablation_sampling_selection},
deficit-driven sampling improves mAP by 3.47 points under
count-based selection and by 3.00 points under deficit-based selection.
Conversely, replacing count-based selection with the marginal-gain
criterion in Eq.~\eqref{eq:marginal_gain} improves mAP by 0.67 points
under natural sampling and by 0.20 points under
deficit-driven sampling.
These results identify deficit-driven object sampling as the main
contributor in this comparison, while marginal-gain selection
provides smaller additional improvements.
\begin{table}[t]
    \centering
    \caption{Sampling and candidate-selection ablations on COCO
at a 0.25\% image budget.
Results are mAP, mean $\pm$ SD over three runs. Bold indicates the best result.}
    \label{tab:ablation_sampling_selection}
    \small
    \setlength{\tabcolsep}{4pt}
    \begin{tabular}{llc}
        \toprule
        Object sampling & Candidate selection & mAP $\uparrow$ \\
        \midrule
        Natural       & Count-based   & $12.13 \pm 0.32$ \\
        Natural       & Deficit-based & $12.80 \pm 0.20$ \\
        Deficit-based & Count-based   & $15.60 \pm 0.20$ \\
        Deficit-based & Deficit-based & $\mathbf{15.80 \pm 0.30}$ \\
        \bottomrule
    \end{tabular}
\end{table}

\paragraph{State feedback and supervision granularity.}
Table~\ref{tab:strategy_comparison} compares sampling
policies under shared target quotas and state-aware
candidate selection.
SGSA (Class--Scale) outperforms static rebalancing
by 0.27 mAP, supporting adaptive deficit-driven sampling.
SGSA (Class) achieves the highest mAP (16.23),
while SGSA (Scale) obtains 14.30.
Compared with SGSA (Class), joint sampling improves
AP$_S$ from 6.83 to 7.13 but yields lower overall mAP,
revealing a performance trade-off.
We adopt joint class--scale sampling as the default
to address cell-level supervision deficits, consistent
with our dataset-level allocation objective.
Variant definitions and full results are provided
in the supplementary material.

\begin{table}[t]
    \centering
    \caption{Ablation of state feedback and supervision granularity on COCO at a 0.25\% image budget. Results are reported as mean $\pm$ standard deviation over three seeds. Bold indicates the main method or the best results.}
    \label{tab:strategy_comparison}
    \small
    \setlength{\tabcolsep}{2pt}
    \resizebox{\columnwidth}{!}{%
    \begin{tabular}{lcccc}
        \toprule
        Variant & mAP & AP$_S$ & AP$_M$ & AP$_L$ \\
        \midrule
        SGSA (Class–Scale)
        & \textbf{15.80 $\pm$ 0.30}
        & \textbf{7.13 $\pm$ 0.15}
        & \textbf{21.63 $\pm$ 0.15}
        & \textbf{21.17 $\pm$ 0.45} \\
        SGSA (Class)
        & \textbf{16.23 $\pm$ 0.12}
        & 6.83 $\pm$ 0.15
        & \textbf{22.57 $\pm$ 0.25}
        & \textbf{21.60 $\pm$ 0.17} \\
        SGSA (Scale)
        & 14.30 $\pm$ 0.36
        & 6.10 $\pm$ 0.36
        & 18.53 $\pm$ 0.31
        & 19.67 $\pm$ 0.40 \\
        Static rebal.
        & 15.53 $\pm$ 0.29
        & 6.77 $\pm$ 0.12
        & 21.17 $\pm$ 0.31
        & 20.87 $\pm$ 0.67 \\
        \bottomrule
    \end{tabular}%
    }
\end{table}

\paragraph{Auxiliary construction components.}
Fixing the recognition-rate estimate to its prior yields
16.03 mAP, compared with 15.80 for full SGSA.
Removing refill($R=0$)  or using $K\in\{1,4\}$ changes mean mAP
by at most 0.17 points.
These small variations suggest that SGSA is robust to
the tested auxiliary configurations.
Complete results are provided in the supplementary material.

\paragraph{Effect of feature distillation.}
We compare SGSA with OD$^3$+gate with and without
PKD, retaining each method's synthesized images
and pseudo-labels while setting $\eta_l=0$ in
Eq.~\eqref{eq:training_loss}.
As shown in Table~\ref{tab:ablation_pkd},
SGSA maintains a 2.47-point mAP advantage
without PKD, positive across all three paired runs,
compared with 3.47 points with PKD.
This confirms that SGSA's gains persist
without feature distillation.

\begin{table}[t]
    \centering
    \caption{Effect of PKD feature distillation.
    Detector scores and paired gains are reported as mean
    $\pm$ SD over three runs.
    The paired gain is SGSA minus OD$^3$+gate.}
    \label{tab:ablation_pkd}
    \small
    \setlength{\tabcolsep}{4pt}
    \begin{tabular}{lcc}
        \toprule
        Method & With PKD & Without PKD \\
        \midrule
        OD$^3$+gate & $12.33 \pm 0.23$ & $6.63 \pm 0.21$ \\
        SGSA        & $15.80 \pm 0.30$ & $9.10 \pm 0.20$ \\
        \midrule
        Paired gain & $+3.47 \pm 0.38$ & $+2.47 \pm 0.15$ \\
        \bottomrule
    \end{tabular}
\end{table}

\subsection{Further Analysis}
\label{sec:further_analysis}

\paragraph{Supervision allocation and quota completion.}
We examine supervision allocation on COCO at a fixed
0.25\% image budget.
Fig.~\ref{fig:mechanism}(a) shows that the five most represented
categories account for 37.6\% of SGSA's counted supervision,
compared with 61.1\% under natural sampling without jitter.
SGSA also achieves higher quota completion, reaching
60.4\% versus 45.8\% (Fig.~\ref{fig:mechanism}(b)). 
Across three construction seeds, 70.2\% of SGSA's counted
supervision contributes to the capped class--scale quotas,
compared with 49.5\% for natural sampling without jitter
and 42.9\% with matched jitter (Fig.~\ref{fig:mechanism}(c)).
These results indicate that SGSA reduces category concentration
while better aligning supervision with the target class--scale
quotas. Further analysis across feedback granularities shows that joint feedback achieves the highest class--scale quota completion (60.4\%), compared with class-only (53.5\%) and scale-only (50.0\%) feedback (see supplementary material). Fig.~\ref{fig:mechanism}(d) further compares the per-canvas counts of placed objects, Observer predictions, and training pseudo-labels, highlighting that the supervision ultimately used for student training depends not only on object placement but also on Observer predictions and pseudo-label filtering.

\paragraph{Generalizability Across Student Detectors.}
We evaluate SGSA against the equal-budget OD$^3$+gate
baseline across student detectors on COCO at a 0.25\%
image budget, reusing each method's synthesized images
and pseudo-labels under matched training settings.
As shown in Table~\ref{tab:student_generalization},
SGSA improves Faster R-CNN R101 by 3.37 mAP points,
comparable to the 3.47-point gain with R50.
For RetinaNet, extending training from 96 to 288 epochs
raises SGSA's mAP from 4.40 to 14.53 and the baseline's
from 1.63 to 9.53, yielding a 5.00-point advantage.
SGSA also outperforms the baseline with FCOS under both
schedules, although absolute mAP remains modest.
\begin{table}[t]
    \centering
    \caption{
    Generalization across student detectors on COCO at a 0.25\%
    image budget.
    Scores are mAP (\%), reported as mean $\pm$ sample standard
    deviation over three paired runs.
    $\Delta$ denotes SGSA minus OD$^3$+gate in mAP points,
    computed before rounding. Bold values indicate the best performance for each setting.
    }
    \label{tab:student_generalization}
    \footnotesize
    \setlength{\tabcolsep}{3pt}
    \renewcommand{\arraystretch}{1.12}
    \begin{tabular}{@{}lcccc@{}}
        \toprule
        Student detector
        & Epochs
        & OD$^3$+gate
        & \textbf{SGSA}
        & $\Delta$ \\
        \midrule
        Faster R-CNN R50
        & 96
        & $12.33 \pm 0.23$
        & $\mathbf{15.80} \pm 0.30$
        & $+3.47$ \\
        Faster R-CNN R101
        & 96
        & $15.77 \pm 0.31$
        & $\mathbf{19.13} \pm 0.31$
        & $+3.37$ \\
        \midrule
        RetinaNet R50
        & 96
        & $1.63 \pm 0.31$
        & $\mathbf{4.40} \pm 0.87$
        & $+2.77$ \\
        RetinaNet R50
        & 288
        & $9.53 \pm 0.67$
        & $\mathbf{14.53} \pm 0.12$
        & $+5.00$ \\
        \midrule
        \addlinespace[2pt]
        FCOS R50
        & 96
        & $0.03 \pm 0.06$
        & $\mathbf{0.10} \pm 0.00$
        & $+0.07$ \\
        FCOS R50
        & 288
        & $0.23 \pm 0.06$
        & $\mathbf{0.43} \pm 0.06$
        & $+0.20$ \\
        \bottomrule
    \end{tabular}
\end{table}

\paragraph{Hyperparameter sensitivity.}
SGSA shows limited sensitivity to most hyperparameters, including the accounting threshold, candidate count, refinement rounds, and target supervision amount. The target-distribution parameter $\alpha$ has a larger impact: compared with the default $\alpha=0.5$, setting $\alpha=1.0$ reduces mAP by 0.70 points, while $\alpha=0.25$ improves it by 0.27 points. Detailed results are provided in the supplementary material.

%% file: sec/5_conclusion.tex
\section{Conclusion}
\label{sec:conclusion}

We presented SGSA, framing object detection dataset distillation as dataset-level progressive supervision allocation. Using accumulated detector feedback, SGSA combines deficit-driven object sampling with marginal-gain candidate selection to promote complementary class--scale supervision under limited image budgets. Experiments on MS COCO and PASCAL VOC show consistent gains across budgets, improved quota utilization, and reduced supervision concentration. Ablations identify deficit-driven sampling as the main contributor, with gains persisting without feature distillation.

%% file: sec/X_suppl.tex
\clearpage

\appendix

\section{Datasets and Evaluation Protocols}
\label{sec:supp_dataset}

We evaluate on MS COCO and PASCAL VOC.
Table~\ref{tab:supp_datasets} summarizes the source splits,
evaluation splits, and nominal image budgets.
Source instance annotations are used to construct the
object pools, while students are trained on the resulting
compact datasets and evaluated on real held-out images.

\paragraph{MS COCO.}
We use COCO 2017 with 80 object categories.
Object instances are extracted from the
\texttt{train2017} annotations, and detection performance
is evaluated on \texttt{val2017}.
For the SGSA object pool, we exclude crowd annotations
and bounding boxes with either side shorter than four pixels.
The retained instances are indexed by category and scale
for dataset construction.
We evaluate at image budgets of 0.25\%, 0.5\%, and 1.0\%.

\paragraph{PASCAL VOC.}
We use the combined VOC2007 and VOC2012
\texttt{trainval} splits as the construction source,
covering 20 object categories.
Evaluation is performed on the VOC2007 \texttt{test} split.
The annotations are represented in COCO JSON format,
with the VOC category metadata explicitly retained
throughout data loading and pseudo-label generation.
We evaluate at image budgets of 0.5\%, 1.0\%, and 2.0\%.

\paragraph{Evaluation metrics.}
For COCO, we report standard bounding-box mAP averaged
over IoU thresholds from 0.50 to 0.95, together with
AP$_{50}$ and AP$_{75}$.
Our VOC experiments use CocoMetric on the converted
VOC2007 test annotations, and the main VOC table reports
mAP$_{50}$, namely category-averaged AP at IoU 0.50.
Additional metrics are reported in the supplementary results.
All detection metrics are expressed as percentages.

\paragraph{Reporting conventions.}
Unless otherwise specified, results from locally
repeated experiments are reported as the mean
and sample standard deviation over three runs.

\begin{table*}[t]
    \centering
    \caption{
        Datasets and nominal image budgets.
        Budget percentages are defined relative to the
        corresponding source training split.
        Construction sources provide the original instances;
        student training uses the constructed compact datasets.
    }
    \label{tab:supp_datasets}
    \small
    \setlength{\tabcolsep}{5pt}
    \renewcommand{\arraystretch}{1.15}
    \begin{tabular}{lclll}
        \toprule
        Dataset & Classes & Construction source
        & Evaluation split & Image budgets \\
        \midrule
        COCO 2017
        & 80
        & train2017
        & val2017
        & 0.25\%, 0.5\%, 1.0\% \\
        PASCAL VOC
        & 20
        & VOC2007/2012 trainval
        & VOC2007 test
        & 0.5\%, 1.0\%, 2.0\% \\
        \bottomrule
    \end{tabular}
\end{table*}

\section{Baseline Details}
\label{sec:supp_baseline_details}

We compare SGSA with real-image subset selection,
model inversion, diffusion-based synthesis,
and object composition baselines.

\paragraph{Subset-selection baselines.}
We include four representative real-image subset-selection
methods: Random, Uniform, K-Center, and Herding.
Random uniformly samples images from the source training
set without considering their semantic content or feature
representations.
Uniform adopts class-balanced sampling to improve the
representation of different object categories in the
selected subset.
K-Center focuses on sample diversity by selecting images
that cover the source feature space. Specifically, it
aims to minimize the maximum feature-space distance
between each source image and its nearest selected sample.
In contrast, Herding emphasizes representativeness by
iteratively selecting images whose aggregated feature
representation approximates the mean feature vector
of the full source dataset.
Unlike SGSA, these methods directly retain selected
real images as the compact training set, without
recomposing object instances onto new canvases.

\paragraph{DCOD.}
DCOD is a detection-specific dataset condensation method
based on a two-stage Fetch-and-Forge pipeline.
It first stores localization and classification
information in detector parameters, then synthesizes
compact training images through model inversion.

\paragraph{UniDD.}
UniDD proposes a task-driven diffusion
framework for dataset distillation across classification,
detection, and segmentation tasks.
It first extracts task-relevant knowledge by training
task-specific proxy models, which subsequently guide
diffusion-based synthesis to generate compact synthetic
datasets with task-aligned supervision.

\paragraph{OD$^3$ and OD$^3$+gate.}
OD$^3$ constructs compact datasets by placing source
object patches on canvases and applying Observer-guided
screening.
On COCO, we reproduce OD$^3$ under image budgets matched
to SGSA, matching the actual number of retained images.
The reproduced OD$^3$ baseline retains its downstream
pseudo-labeling pipeline without our additional final
confidence filter.

OD$^3$+gate applies the same final pseudo-label threshold
$\gamma_{\mathrm{label}}=0.3$ used by SGSA.
This filter acts on the training annotations after
pseudo-label generation; it does not introduce SGSA's
deficit-driven construction policy.
The local OD$^3$ variants and SGSA use matched student
architectures, initialization, training schedules,
and feature-distillation settings.

\section{Implementation Details}
\label{sec:supp_implementation}

We report the configurations used in our experiments.
Unless otherwise specified, construction and training
hyperparameters are shared across MS COCO and PASCAL VOC.

\subsection{Dataset Construction}
\paragraph{Object pool construction.}
We construct the category--scale object pools
$\{\mathcal{P}_{c,s}\}$ from the source instance
annotations of each dataset.
Crowd annotations and boxes with either side
shorter than four pixels are removed.
Each retained instance is indexed by category
$c$ and scale bin $s\in\{0,1,2\}$, using
thresholds $\tau_1=32$ and $\tau_2=96$.
Object area is measured after the reference-canvas
rescaling in Eq.~\eqref{eq:reference_scaling},
which preserves the object-to-image area ratio.

On COCO, the pool contains $838{,}912$ instances
from $118{,}287$ images across 80 categories,
including $251{,}622$ small, $299{,}800$ medium,
and $287{,}490$ large objects.
On VOC, we combine the VOC2007 and VOC2012
\texttt{trainval} splits, obtaining $40{,}058$
instances from $16{,}551$ images across 20
categories.
The VOC pool contains $1{,}187$ small,
$9{,}977$ medium, and $28{,}894$ large objects.
Small objects thus account for approximately
$3\%$ of VOC instances, compared with $30\%$
on COCO.
Each pool entry records the source image identifier,
category, source box, and canvas-rescaled size.
Object pixels are read from source images during
construction without caching cropped patches.

\paragraph{Canvas composition.}

All canvases have a resolution of
$W\times H=484\times578$ pixels, matching
the OD$^3$ canvas geometry.
For both datasets, backgrounds are selected
from ranked lists of the corresponding source
images: COCO train2017 ($118{,}287$ images)
and the combined VOC2007 and VOC2012
\texttt{trainval} splits ($16{,}551$ images).
All pasted objects are cropped from real
source images, and no generative model is used.

Object patches are extracted with size-adaptive
context margins to preserve surrounding
visual information. Each patch undergoes independent isotropic scale
jitter $j_i\sim\mathcal{U}(j_{\min},1)$,
with $j_{\min}=0.5$.
Larger patches are assigned to the back layer
and smaller patches to the front layer according
to their jittered areas.
Within each layer, patches are placed in
descending area order.
Each placement attempt considers up to $40$
random positions, with occupancy thresholds
$\tau=0.6$ for front-layer objects and
$\tau_{\mathrm{back}}=0.3$ among large objects
on the back layer.
The minimum visible fraction for back-layer
objects is $0.5$.
The planner permits at most three large-scale
object selections per canvas, and each object
receives at most two placement attempts.
The planner proposes up to $120$ objects per
candidate.
All composition parameters are shared across
COCO and VOC.

For COCO, image budgets of $0.25\%$,
$0.5\%$, and $1.0\%$ correspond to
$296$, $592$, and $1{,}184$ synthesized
canvases, respectively.
For VOC, budgets of $0.5\%$, $1.0\%$,
and $2.0\%$ correspond to $83$, $166$,
and $331$ canvases, respectively.

\subsection{Feedback-Guided Construction}
\paragraph{Target quota and global state.}
The target allocation in Eq.~\eqref{eq:target}
uses a per-canvas supervision quota
$\lambda=50$ and class-tempering exponent
$\alpha=0.5$.
The category distribution $p_c$ and conditional
scale distribution $p_{s\mid c}$ are computed
from the corresponding source object pool. 

The global state $S_t$ accumulates the
Observer-predicted category--scale histograms
$H(I)$ over accepted canvases.
The empirical recognition rate in
Eq.~\eqref{eq:recognition_rate} uses prior
$p_0=0.3$ and smoothing constant $\kappa=4$.
Successful recognition requires an Observer
prediction with the same category,
confidence at least $\gamma_{\mathrm{rec}}=0.3$, and
IoU at least $0.5$ with the transformed
source box.
The numerical stability constant in
Eq.~\eqref{eq:sampling_weights}
is set to $\epsilon=10^{-9}$.

\paragraph{Observer feedback and candidate selection.}
The Observer is a pre-trained Faster R-CNN detector.
COCO uses the R101-FPN checkpoint
\textit{faster\_rcnn\_r101\_fpn\_2x\_coco},
while VOC uses the R50-FPN checkpoint
\textit{faster\_rcnn\_r50\_fpn\_1x\_voc0712}.
The corresponding Observer checkpoint is also
used as the distillation teacher.

For each dataset position, we construct $K=2$
candidate canvases.
Each candidate undergoes $R=2$ refill rounds,
with three filtering passes interleaved with
two refill steps.
During filtering, a pasted object is retained
if its transformed source box overlaps any
Observer prediction with IoU greater than $0.25$,
without requiring category agreement or an
additional confidence threshold. The retention criterion is intentionally permissive,
whereas recognition-rate estimation requires stricter
category-consistent matching to avoid overestimating
successful recognition.
A final detection pass produces the histogram
$H(I)$ used for candidate scoring and
supervision accounting.

Predictions with confidence $q_j \geq \gamma_{\mathrm{gain}}=0.3$
are counted for supervision accounting.
Candidates are scored using the deficit-capped
marginal gain in Eq.~\eqref{eq:marginal_gain}.
The highest-scoring candidate is accepted,
and the global state is updated according
to Eq.~\eqref{eq:state_update}.
The same selection parameters are used on
COCO and VOC.

\subsection{Pseudo-Labeling and Student Training}

\paragraph{Pseudo-label generation.}
During construction, single-pass Observer
predictions provide the histograms $H(I)$
used for candidate scoring and state updates.
After construction, all accepted canvases are
re-annotated using the same detector with
test-time augmentation (TTA) at scales
$\{(320,320),(640,640),(960,960)\}$ and
horizontal flipping.
Predictions are merged by NMS at IoU $0.6$,
restricted to the top $100$ detections per image,
and filtered at
$\gamma_{\mathrm{label}}=0.3$.

Each pseudo-label contains a bounding box,
a hard category label, and a confidence score.
Only boxes and categories are used for
student training.
At the $0.25\%$ COCO budget, the final filtering
retains $18{,}720$ of $29{,}563$ detections.
VOC uses the same pseudo-labeling procedure,
with annotations stored in COCO JSON format
and the VOC 20-category metadata specified
for all data loaders.

\paragraph{Student training and distillation.}
The student is Faster R-CNN R50-FPN, trained
for $96$ epochs using SGD with momentum $0.9$
and weight decay $10^{-4}$.
Training uses two GPUs with batch size two per
GPU.
A linear warm-up runs for $500$ iterations
with start factor $10^{-3}$.
The learning rate is multiplied by $0.1$ at
epochs $64$ and $88$, with validation every
four epochs.

The initial learning rate is $0.02$ for COCO
and $0.0025$ for VOC.
The VOC rate follows the MMDetection reference
setting of $0.01$ at batch size $16$, scaled
to our total batch size of four.
All other training parameters are identical.

Feature distillation uses PKD on four FPN
levels ($L=4$), indexed by
$l\in\{0,1,2,3\}$, with a uniform loss
weight $\eta_l=6$ per level.
The student backbone is initialized with
ImageNet-pretrained ResNet-50 weights.
The same initialization is used for all
compared methods.

\subsection{Reproducibility and hardware.}
We use three paired synthesis and training
seeds, $i\in\{0,1,2\}$.
Each seed produces an independently synthesized
dataset and a corresponding student training run.
Results are reported as means and sample
standard deviations over the three runs.
Ablation differences are computed per seed
before averaging, with sign consistency also
recorded.

Experiments run on NVIDIA GeForce RTX 3090 GPUs
under SLURM, using one GPU for synthesis and
two GPUs for student training.
Completed runs are checked for finite losses,
valid checkpoints and metrics, and consistency
between configured and recorded dataset sizes.

\section{Additional Ablation Studies}
\label{sec:supp_additional_ablations}

We conduct additional ablations on COCO at a fixed image
budget of 0.25\%, using a Faster R-CNN R50 student trained
for 96 epochs.
Unless otherwise specified, pseudo-label filtering and
student training follow the main experimental protocol.
We report mean mAP and sample standard deviation over three
paired runs, with synthesis and training seeds matched
within each run.

\subsection{Auxiliary Construction Ablations}
We examine recognition-rate estimation, refill, candidate
count, placement geometry, and supervision accounting.
For recognition-rate estimation, we replace the online
estimate $\rho(c,s)$ with the fixed prior $p_0=0.3$.
For refill, we set $R=0$ while retaining the initial
Observer filtering.
We additionally vary the candidate count, apply the
OD$^3$ object-size cap within SGSA, and replace the layered
layout with a single-layer layout.
Finally, we restrict the prediction counts used for
candidate scoring and state updates to same-class,
one-to-one matches with pasted objects at IoU $\geq 0.5$.
This restriction changes construction-stage accounting,
while retaining the downstream pseudo-labeling protocol.

As shown in Table~\ref{tab:supp_auxiliary}, these variants
change mean mAP by between $-0.20$ and $+0.23$ points.
The fixed-prior variant obtains 16.03 mAP, compared with
15.80 for full SGSA, providing no observed accuracy
advantage for online recognition-rate updates.
One-to-one accounting yields 15.90 mAP, indicating similar
performance under this stricter accounting rule.
Overall, the small mean variations suggest robustness
to the tested auxiliary configurations.

\begin{table}[t]
    \centering
    \caption{
        Additional construction ablations on COCO at a
        0.25\% image budget.
        Results are mean mAP $\pm$ sample SD over three runs.
        Full SGSA uses online recognition-rate updates,
        $R=2$, and $K=2$.
        $\Delta$ denotes the mean paired mAP difference
        from full SGSA, in points.
    }
    \label{tab:supp_auxiliary}
    \small
    \setlength{\tabcolsep}{4pt}
    \renewcommand{\arraystretch}{1.08}
    \begin{tabular}{lcc}
        \toprule
        Variant & mAP (\%) $\uparrow$ & $\Delta$ \\
        \midrule
        Full SGSA
            & $15.80 \pm 0.30$ & --- \\
        Fixed $\rho=p_0$
            & $16.03 \pm 0.21$ & $+0.23$ \\
        No refill ($R=0$)
            & $15.93 \pm 0.15$ & $+0.13$ \\
        One candidate ($K=1$)
            & $15.63 \pm 0.32$ & $-0.17$ \\
        Four candidates ($K=4$)
            & $15.63 \pm 0.12$ & $-0.17$ \\
        \midrule
        OD$^3$ object-size cap
            & $15.67 \pm 0.23$ & $-0.13$ \\
        Single-layer layout
            & $15.60 \pm 0.26$ & $-0.20$ \\
        One-to-one accounting
            & $15.90 \pm 0.17$ & $+0.10$ \\
        \bottomrule
    \end{tabular}
\end{table}

\subsection{Sampling Feedback and Supervision Granularity}
\paragraph{Definition of the feedback variants.}
The four variants in Table~\ref{tab:strategy_per_seed}
share the same target allocation $T(c,s)$, source object
pools, Observer, canvas geometry, and hyperparameters:
$\alpha=0.5$, $\lambda=50$, $\gamma=0.3$, $K=2$,
$R=2$, and scale jitter in $[0.5,1.0]$.
They differ in the object-sampling rule used to select
the next class--scale pool.
Canvas composition, Observer filtering, joint marginal-gain
candidate selection, the state-update rule, pseudo-labeling,
and student training are kept unchanged.
In particular, static rebalancing refers only to the
sampling rule and retains the same state-aware candidate
selection as the SGSA variants.

Let $S_t(c,s)$ denote the accumulated construction-stage
supervision after accepting $t$ canvases.
Within a planning call, let $P^{(\ell)}(c,s)$ count the
objects selected from pool $(c,s)$ before sampling step
$\ell$, and let $\rho_t(c,s)$ denote the current empirical
recognition-rate estimate.
The planning counts are reset at the beginning of each
call, including refill calls.
We write $[x]_+=\max(x,0)$ and define the estimated
contribution of the planned objects as
\[
E_t^{(\ell)}(c,s)
=
P^{(\ell)}(c,s)\rho_t(c,s).
\]
The corresponding marginal quantities are
\[
\begin{aligned}
T_c &= \sum_s T(c,s),
& T_s &= \sum_c T(c,s),\\
S_{t,c} &= \sum_s S_t(c,s),
& S_{t,s} &= \sum_c S_t(c,s),\\
E_{t,c}^{(\ell)} &= \sum_s E_t^{(\ell)}(c,s),
& E_{t,s}^{(\ell)} &= \sum_c E_t^{(\ell)}(c,s).
\end{aligned}
\]
We specify the unmasked sampling weights
$\widetilde{w}$ for each variant below.

\begin{description}

\item[SGSA (Class--Scale).]
The full SGSA method computes the remaining supervision
deficit for each class--scale cell:
\[
\widetilde{w}_{t,\mathrm{joint}}^{(\ell)}(c,s)
=
\frac{
\left[T(c,s)-S_t(c,s)-E_t^{(\ell)}(c,s)\right]_+
}{
\max(T(c,s),\epsilon)
}
\rho_t(c,s).
\]
The cumulative deficit is evaluated on the joint
$C\times 3$ supervision state, with $C=80$ on COCO.
This allows sampling to prioritize specific
class--scale combinations that remain under-covered.

\item[SGSA (Scale).]
This variant aggregates the supervision deficit
over classes within each scale bin:
\[
D_{t,s}^{(\ell)}
=
\left[T_s-S_{t,s}-E_{t,s}^{(\ell)}\right]_+.
\]
The sampling weights are
\[
\widetilde{w}_{t,\mathrm{scale}}^{(\ell)}(c,s)
=
\frac{D_{t,s}^{(\ell)}}{\max(T_s,\epsilon)}
\frac{T(c,s)}{\max(T_s,\epsilon)}
\rho_t(c,s).
\]
The cumulative deficit is scale-level, while
the target-conditional class distribution and
the per-pool recognition rate determine the
allocation within each scale bin.

\item[SGSA (Class).]
This variant aggregates the supervision deficit
over scales within each class:
\[
D_{t,c}^{(\ell)}
=
\left[T_c-S_{t,c}-E_{t,c}^{(\ell)}\right]_+.
\]
The sampling weights are
\[
\widetilde{w}_{t,\mathrm{class}}^{(\ell)}(c,s)
=
\frac{D_{t,c}^{(\ell)}}{\max(T_c,\epsilon)}
\frac{T(c,s)}{\max(T_c,\epsilon)}
\rho_t(c,s).
\]
The cumulative deficit is class-level, while
the target-conditional scale distribution and
the per-pool recognition rate determine the
allocation within each class.

\item[Static rebalancing.]
This variant uses fixed target-proportional
sampling weights:
\[
\widetilde{w}_{t,\mathrm{static}}^{(\ell)}(c,s)
=
T(c,s).
\]
The sampling rule does not use the accumulated
state $S_t$ or the recognition-rate estimate
$\rho_t$.
However, the shared candidate-selection stage
remains state-dependent.

\end{description}

For every variant $a$, the same masking rules
enforce pool availability and the per-canvas
large-object limit:
\[
w_{t,a}^{(\ell)}(c,s)
=
M^{(\ell)}(c,s)
\widetilde{w}_{t,a}^{(\ell)}(c,s).
\]
When the total sampling weight is positive,
the sampling distribution is
\[
\pi_{t,a}^{(\ell)}(c,s)
=
\frac{
w_{t,a}^{(\ell)}(c,s)
}{
\sum_{c',s'}w_{t,a}^{(\ell)}(c',s')
}.
\]

SGSA (Class--Scale), SGSA (Scale), and SGSA (Class)
differ in the granularity at which accumulated
supervision deficits are computed during sampling.
All three variants retain joint class--scale
candidate selection.
Static rebalancing instead uses fixed
target-proportional sampling while retaining
the same state-aware candidate-selection rule.
Thus, this ablation compares alternative
sampling policies under shared target quotas
and construction settings, without changing
the downstream training protocol.

\paragraph{Per-run feedback comparisons.}
Table~\ref{tab:strategy_per_seed} reports the
per-seed detection results of the four sampling
variants under matched construction and training
settings.
SGSA (Class--Scale) consistently outperforms
static rebalancing across the three paired runs,
with gains of 0.40, 0.10, and 0.30 mAP points,
respectively, yielding an average improvement
of 0.27 points.
Since both variants retain state-aware candidate
selection, this comparison demonstrates the
additional benefit of deficit-driven sampling
over fixed target-proportional sampling.

SGSA (Class--Scale) also outperforms SGSA (Scale)
by 1.50 mAP points on average, suggesting the
importance of class-conditioned supervision
deficits during sampling.
Although SGSA (Class) achieves the highest
overall mAP (16.23), SGSA (Class--Scale)
improves $\mathrm{AP}_S$ from 6.83 to 7.13.
These results reveal a trade-off between overall
detection accuracy and small-object performance
when incorporating joint class--scale feedback.

\begin{table}[t]
    \centering
    \caption{Per-seed results of different state feedback and supervision granularity strategies on COCO at a 0.25\% image budget. All variants are evaluated over three random seeds.}
    \label{tab:strategy_per_seed}
    \small
    \setlength{\tabcolsep}{5pt}
    \begin{tabular}{lccccc}
        \toprule
        Variant & Seed & mAP & AP$_S$ & AP$_M$ & AP$_L$ \\
        \midrule
        SGSA (Class–Scale) & 0 & 16.10 & 7.10 & 21.80 & 21.60 \\
                          & 1 & 15.80 & 7.30 & 21.60 & 21.20 \\
                          & 2 & 15.50 & 7.00 & 21.50 & 20.70 \\
        \midrule
        SGSA (Class)        & 0 & 16.10 & 6.80 & 22.30 & 21.40 \\
                          & 1 & 16.30 & 7.00 & 22.80 & 21.70 \\
                          & 2 & 16.30 & 6.70 & 22.60 & 21.70 \\
        \midrule
        SGSA (scale)       & 0 & 14.40 & 6.20 & 18.80 & 19.60 \\
                          & 1 & 14.60 & 6.40 & 18.60 & 20.10 \\
                          & 2 & 13.90 & 5.70 & 18.20 & 19.30 \\
        \midrule
        Static rebal.     & 0 & 15.70 & 6.70 & 21.10 & 21.30 \\
                          & 1 & 15.70 & 6.90 & 21.50 & 21.20 \\
                          & 2 & 15.20 & 6.70 & 20.90 & 20.10 \\
        \bottomrule
    \end{tabular}
\end{table}

\begin{table}[t]
\centering
\small
\setlength{\tabcolsep}{4pt}
\caption{
Construction-stage quota completion (\%) for three
sampling-feedback granularities, evaluated against
the same target quotas.
All variants retain joint marginal-gain candidate
selection and the same state-update rule.
Bold indicates the highest reported completion
within each column.
}
\label{tab:supp_feedback_coverage}
\begin{tabular}{lccc}
\toprule
Sampling feedback & Class & Scale & Joint \\
\midrule
SGSA(Class--scale) & 68.9 & 66.6 & \textbf{60.4} \\
SGSA(Class)        & \textbf{75.2} & 59.7 & 53.5 \\
SGSA(Scale)        & 56.0 & \textbf{72.9} & 50.0 \\
\bottomrule
\end{tabular}
\end{table}

\paragraph{Effect of feedback granularity.}
Table~\ref{tab:supp_feedback_coverage} shows that
joint class--scale sampling achieves the highest
joint quota completion among these three variants:
60.4\%, compared with 53.5\% for class-only sampling
and 50.0\% for scale-only sampling.
Class-only sampling attains the highest class-level
completion (75.2\%), while scale-only sampling attains
the highest scale-level completion (72.9\%).
Relative to joint sampling, each marginal variant
achieves higher completion along its corresponding
axis but lower completion along the other axis.

Together with the detector results in
Table~\ref{tab:strategy_per_seed}, these measurements
characterize a trade-off between joint supervision
coverage and overall detection accuracy.
Joint sampling achieves higher joint quota completion
and small-object AP, whereas class-only sampling
yields higher overall mAP.

\begin{figure*}[t]
    \centering
    \includegraphics[width=\textwidth]
      {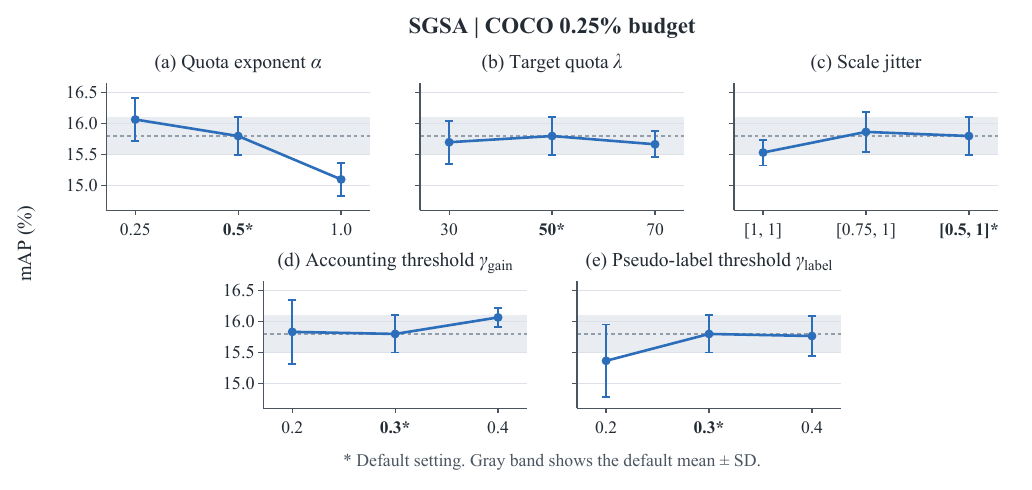}
    \caption{
    Hyperparameter sensitivity of SGSA on COCO
    at a 0.25\% image budget.
    Five hyperparameters are varied individually,
    with all other settings held fixed.
    Points and error bars indicate mean mAP and
    sample standard deviation over three paired runs.
    Asterisks mark the default settings, and
    gray shading denotes the default mean
    $\pm$ one sample standard deviation.
}
    \label{fig:supp_hyperparameters}
\end{figure*}

\section{Hyperparameter Sensitivity}
\label{sec:supp_hyperparameter_sensitivity}

We investigate the sensitivity of SGSA to five
hyperparameters on COCO at a 0.25\% image budget,
varying one parameter at a time under the same
training protocol.
Figure~\ref{fig:supp_hyperparameters} visualizes
the performance trends, while
Table~\ref{tab:supp_hyperparameters} provides
the complete numerical results.
We report mean mAP and sample standard deviation
over three paired runs.
Default configurations reuse the corresponding
main experimental results.

\paragraph{Target quota parameters.}
The class-frequency exponent $\alpha$ controls
the target quota allocation across categories,
while $\lambda$ determines its overall scale.
As shown in Fig.~\ref{fig:supp_hyperparameters}(a),
setting $\alpha=0.25$ achieves $16.07 \pm 0.35$
mAP, compared with $15.80 \pm 0.30$ at the
default $\alpha=0.5$ and $15.10 \pm 0.26$
at $\alpha=1.0$.
This indicates that class-aware quota allocation
has a noticeable influence on detection performance.
In contrast, varying $\lambda$ from 30 to 70
results in relatively small changes, with mean
mAP ranging from 15.67 to 15.80
(Fig.~\ref{fig:supp_hyperparameters}(b)).

\paragraph{Scale jitter.}
Figure~\ref{fig:supp_hyperparameters}(c) compares
three scale-jitter intervals.
Without scale jitter ($[1,1]$), the mean mAP
is 15.53, while intervals of $[0.75,1]$ and
$[0.5,1]$ yield 15.87 and 15.80, respectively.
These results show that SGSA maintains
comparable detection performance across the
tested scale-jitter configurations.

\begin{table}[t]
    \centering
    \caption{
        Complete parameter sweeps on COCO at a 0.25\%
        image budget.
        Results are mean mAP $\pm$ sample SD over three
        paired runs.
        Asterisks denote default settings.
        $\Delta$ is the mean paired mAP difference from
        the default configuration, in points.
    }
    \label{tab:supp_hyperparameters}
    \small
    \setlength{\tabcolsep}{4pt}
    \renewcommand{\arraystretch}{1.06}
    \begin{tabular}{llcc}
        \toprule
        Parameter & Value & mAP (\%) $\uparrow$ & $\Delta$ \\
        \midrule
        $\alpha$
            & $0.25$ & $16.07 \pm 0.35$ & $+0.27$ \\
            & $0.5^{*}$ & $15.80 \pm 0.30$ & --- \\
            & $1.0$ & $15.10 \pm 0.26$ & $-0.70$ \\
        \midrule
        $\gamma_{\mathrm{gain}}$
            & $0.2$ & $15.83 \pm 0.51$ & $+0.03$ \\
            & $0.3^{*}$ & $15.80 \pm 0.30$ & --- \\
            & $0.4$ & $16.07 \pm 0.15$ & $+0.27$ \\
        \midrule
        $\gamma_{\mathrm{label}}$
            & $0.2$ & $15.37 \pm 0.59$ & $-0.43$ \\
            & $0.3^{*}$ & $15.80 \pm 0.30$ & --- \\
            & $0.4$ & $15.77 \pm 0.32$ & $-0.03$ \\
        \midrule
        Scale jitter
            & $[1,1]$ & $15.53 \pm 0.21$ & $-0.27$ \\
            & $[0.75,1]$ & $15.87 \pm 0.32$ & $+0.07$ \\
            & $[0.5,1]^{*}$ & $15.80 \pm 0.30$ & --- \\
        \midrule
        $\lambda$
            & $30$ & $15.70 \pm 0.35$ & $-0.10$ \\
            & $50^{*}$ & $15.80 \pm 0.30$ & --- \\
            & $70$ & $15.67 \pm 0.21$ & $-0.13$ \\
        \bottomrule
    \end{tabular}
\end{table}

\paragraph{Confidence thresholds.}
The accounting threshold $\gamma_{\mathrm{gain}}$
determines which predictions contribute to
candidate scoring and state updates.
The recognition-verification threshold remains
fixed at 0.3 throughout this sweep.
As shown in Fig.~\ref{fig:supp_hyperparameters}(d),
varying $\gamma_{\mathrm{gain}}$ from 0.2 to 0.4
produces mean mAP values between 15.80 and 16.07,
indicating limited sensitivity to this threshold.

For the final pseudo-label threshold
$\gamma_{\mathrm{label}}$, we reuse the same
constructed images and unfiltered predictions
for each seed, varying only the final confidence
filter.
Thresholds of 0.2, 0.3, and 0.4 yield
$15.37 \pm 0.59$, $15.80 \pm 0.30$, and
$15.77 \pm 0.32$ mAP, respectively
(Fig.~\ref{fig:supp_hyperparameters}(e)).
The lower threshold of 0.2 results in reduced
mean accuracy and greater variation across runs,
while the default threshold of 0.3 achieves
competitive performance with relatively
low variability.

\paragraph{Overall observations.}
SGSA exhibits stable detection performance
across most tested hyperparameter configurations.
Changes in the target quota scale, accounting
threshold, and scale-jitter interval result
in mean mAP differences of at most 0.27 points
from the default settings.
The class-frequency exponent has a more
pronounced influence, highlighting the
importance of class-aware target allocation.
Overall, these results support the robustness
of SGSA to most hyperparameter choices,
reducing its dependence on extensive
parameter tuning.

\section{Qualitative Analysis of Constructed Datasets}
\label{sec:supp_qualitative_case_studies}

\begin{figure*}[t]
    \centering
    \includegraphics[width=0.95\textwidth]
        {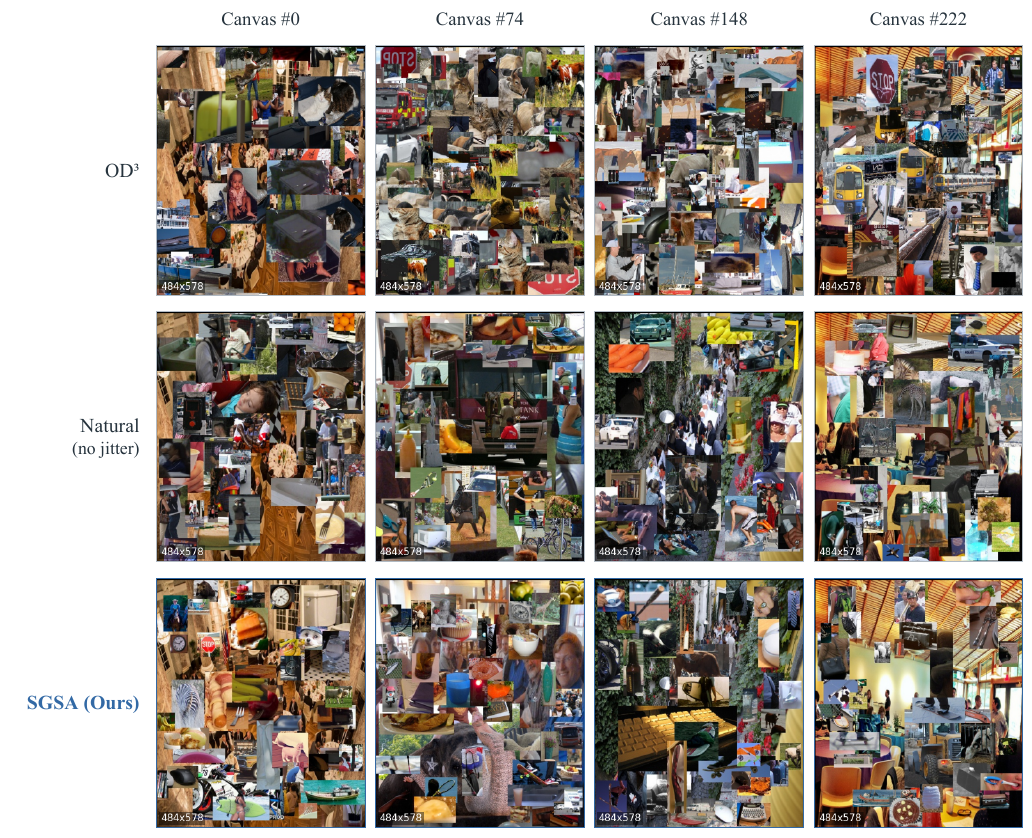}
    \caption{
        Qualitative comparison of constructed COCO
        canvases at the nominal 0.25\% image budget.
        Rows show OD$^3$, Natural, and SGSA,
        all using seed 0.
        Columns correspond to output indices
        0, 74, 148, and 222; selected object
        instances need not match across methods.
        Natural uses natural object sampling and
        count-based candidate selection without
        scale jitter.
        SGSA uses deficit-driven sampling and
        selection with scale jitter $[0.5,1.0]$,
        combining heterogeneous object patches
        at different sizes.
        The OD$^3$ examples come from its original
        synthesis run at this nominal budget.
        Images are shown without detection overlays.
    }
    \label{fig:supp_qualitative}
\end{figure*}

\paragraph{Comparison setup.}
We compare COCO canvases constructed by OD$^3$,
the natural-sampling variant (Natural), and SGSA
at the nominal 0.25\% image budget.
Figure~\ref{fig:supp_qualitative} presents seed-0
examples at four fixed output indices:
0, 74, 148, and 222.
The indices are shared across methods, but the
selected object instances need not match.
Natural uses no scale jitter, whereas SGSA
uses its default interval $[0.5,1.0]$.

\paragraph{Qualitative comparison.}
The SGSA examples illustrate its ability to
combine heterogeneous object patches at different
sizes within compact training canvases.
For example, the SGSA canvas at index 74 combines
person patches with cups, bowls, and food-related
patches at different relative sizes.
The other SGSA examples likewise bring together
multiple object types within individual canvases.
Scale jitter provides additional flexibility
in the sizes of pasted objects, complementing
the diversity of the selected source instances.

The key advantage of SGSA lies not simply in
producing dense compositions, but in directing
their construction toward unmet supervision needs.
OD$^3$ applies Observer-guided screening, while
Natural uses natural object sampling and
count-based candidate selection.
SGSA instead uses the remaining class-scale
supervision deficits to guide both object
sampling and candidate selection.
Consequently, candidates are assessed not merely
by how many high-confidence predictions they
produce, but by how those predictions contribute
to the target supervision allocation.

This distinction is particularly relevant under
a compact image budget: a canvas with many
predictions need not supply the class-scale
supervision that is still missing.
By adapting construction to the accumulated
Observer feedback, SGSA can prioritize groups
that remain below their target quotas.
Together, deficit-driven allocation and scale
jitter provide targeted control over supervision
composition and flexibility in object size,
rather than relying on natural sampling and
prediction counts alone.

\begin{table}[t]
    \centering
    \caption{
        Per-canvas construction statistics for
        the Natural and SGSA examples in
        Fig.~\ref{fig:supp_qualitative}.
        Placed denotes retained pasted objects;
        Counted denotes high-confidence Observer
        predictions used for construction-stage
        accounting, rather than final training
        pseudo-labels.
    }
    \label{tab:supp_qualitative_counts}
    \small
    \setlength{\tabcolsep}{4pt}
    \renewcommand{\arraystretch}{1.08}
    \begin{tabular}{ccccc}
        \toprule
        Canvas index
        & \multicolumn{2}{c}{Natural}
        & \multicolumn{2}{c}{SGSA} \\
        \cmidrule(lr){2-3}\cmidrule(l){4-5}
        & Placed & Counted & Placed & Counted \\
        \midrule
        0   & 37 & 53 & 42 & 43 \\
        74  & 39 & 40 & 44 & 48 \\
        148 & 25 & 51 & 33 & 42 \\
        222 & 39 & 52 & 33 & 54 \\
        \bottomrule
    \end{tabular}
\end{table}

\paragraph{Canvas-level analysis.}
Table~\ref{tab:supp_qualitative_counts} reports
the construction statistics of the displayed
Natural and SGSA canvases.
\emph{Placed} denotes the number of pasted
objects retained on a canvas, whereas
\emph{Counted} denotes the high-confidence
Observer predictions used for construction-stage
accounting.
These predictions are distinct from the final
pseudo-labels used for student training.

SGSA retains more objects than Natural in three
of the four displayed canvases, illustrating
that deficit-driven construction is compatible
with rich multi-object compositions.
At index 74, for example, SGSA retains 44 objects
and counts 48 predictions, compared with 39
objects and 40 predictions for Natural.
However, the two counts do not always increase
together.
At index 0, SGSA retains more objects than
Natural (42 versus 37), but counts fewer
predictions (43 versus 53).
At index 222, the relationship is reversed:
SGSA retains fewer objects (33 versus 39),
but counts more predictions (54 versus 52).

These cases distinguish the physical content
of a canvas from the supervision observed
during construction.
Both Natural and SGSA use Observer predictions,
but SGSA further evaluates their contribution
to the remaining class-scale deficits.
Its construction criterion therefore goes beyond
maximizing either pasted-object counts or
prediction totals, explicitly favoring
supervision that addresses unmet target quotas.

Overall, these examples illustrate how SGSA
combines flexible multi-object composition
with a supervision-aware construction objective.
Its central design advantage is explicit control
over which class-scale supervision to prioritize
within a limited image budget.
Dataset-level coverage and downstream detection
accuracy are assessed by the quantitative
experiments.

%% file: main.bib
@String(CVPR= {IEEE Conf. Comput. Vis. Pattern Recog.})

@String(ICCV= {Int. Conf. Comput. Vis.})

@String(AAAI = {AAAI})

@String(CVPRW= {IEEE Conf. Comput. Vis. Pattern Recog. Worksh.})

@String(CVPR  = {CVPR})

@String(ICCV  = {ICCV})

@String(CVPRW= {CVPRW})

@article{wang2018dataset,
  title={Dataset distillation},
  author={Wang, Tongzhou and Zhu, Jun-Yan and Torralba, Antonio and Efros, Alexei A},
  journal={arXiv preprint arXiv:1811.10959},
  year={2018}
}

@article{zhao2020dataset,
  title={Dataset condensation with gradient matching},
  author={Zhao, Bo and Mopuri, Konda Reddy and Bilen, Hakan},
  journal={arXiv preprint arXiv:2006.05929},
  year={2020}
}

@inproceedings{zhao2021dataset,
  title={Dataset condensation with differentiable siamese augmentation},
  author={Zhao, Bo and Bilen, Hakan},
  booktitle={International Conference on Machine Learning},
  pages={12674--12685},
  year={2021},
  organization={PMLR}
}

@inproceedings{cazenavette2022dataset,
  title={Dataset distillation by matching training trajectories},
  author={Cazenavette, George and Wang, Tongzhou and Torralba, Antonio and Efros, Alexei A and Zhu, Jun-Yan},
  booktitle={2022 IEEE/CVF Conference on Computer Vision and Pattern Recognition (CVPR)},
  pages={10708--10717},
  year={2022},
  organization={IEEE}
}

@inproceedings{zhao2023dataset,
  title={Dataset condensation with distribution matching},
  author={Zhao, Bo and Bilen, Hakan},
  booktitle={2023 IEEE/CVF Winter Conference on Applications of Computer Vision (WACV)},
  pages={6503--6512},
  year={2023},
  organization={IEEE}
}

@inproceedings{zhao2023improved,
  title={Improved distribution matching for dataset condensation},
  author={Zhao, Ganlong and Li, Guanbin and Qin, Yipeng and Yu, Yizhou},
  booktitle={2023 IEEE/CVF Conference on Computer Vision and Pattern Recognition (CVPR)},
  pages={7856--7865},
  year={2023},
  organization={IEEE}
}

@inproceedings{hu2026focusdc,
  title={FocusDC: Real-World Scene Infusion for Robust Dataset Condensation},
  author={Hu, Youbing and Cheng, Yun and Saukh, Olga and Ozdemir, Firat and Lu, Anqi and Cao, Zhiqiang and Zhang, Min and Li, Zhijun},
  booktitle={The Third Conference on Parsimony and Learning (Proceedings Track)},
  year={2026}
}

@article{qi2024fetch,
  title={Fetch and forge: Efficient dataset condensation for object detection},
  author={Qi, Ding and Li, Jian and Peng, Jinlong and Zhao, Bo and Dou, Shuguang and Li, Jialin and Zhang, Jiangning and Wang, Yabiao and Wang, Chengjie and Zhao, Cairong},
  journal={Advances in Neural Information Processing Systems},
  volume={37},
  pages={119283--119300},
  year={2024}
}

@inproceedings{al2026od,
  title={{OD}$^{3}$: Optimization-free Dataset Distillation for Object Detection},
  author={Al Khatib, Salwa and Elhagry, Ahmed and Shao, Shitong and Shen, Zhiqiang},
  booktitle={International Conference on Learning Representations},
  volume={2026},
  pages={104108--104125},
  year={2026}
}

@article{sener2017active,
  title={Active learning for convolutional neural networks: A core-set approach},
  author={Sener, Ozan and Savarese, Silvio},
  journal={arXiv preprint arXiv:1708.00489},
  year={2017}
}

@inproceedings{mirzasoleiman2020coresets,
  title={Coresets for data-efficient training of machine learning models},
  author={Mirzasoleiman, Baharan and Bilmes, Jeff and Leskovec, Jure},
  booktitle={International Conference on Machine Learning},
  pages={6950--6960},
  year={2020},
  organization={PMLR}
}

@inproceedings{killamsetty2021grad,
  title={Grad-match: Gradient matching based data subset selection for efficient deep model training},
  author={Killamsetty, Krishnateja and Durga, Sivasubramanian and Ramakrishnan, Ganesh and De, Abir and Iyer, Rishabh},
  booktitle={International Conference on Machine Learning},
  pages={5464--5474},
  year={2021},
  organization={PMLR}
}

@article{ash2019deep,
  title={Deep batch active learning by diverse, uncertain gradient lower bounds},
  author={Ash, Jordan T and Zhang, Chicheng and Krishnamurthy, Akshay and Langford, John and Agarwal, Alekh},
  journal={arXiv preprint arXiv:1906.03671},
  year={2019}
}

@inproceedings{yu2022consistency,
  title={Consistency-based active learning for object detection},
  author={Yu, Weiping and Zhu, Sijie and Yang, Taojiannan and Chen, Chen},
  booktitle={2022 IEEE/CVF Conference on Computer Vision and Pattern Recognition Workshops (CVPRW)},
  pages={3950--3959},
  year={2022},
  organization={IEEE}
}

@inproceedings{lee2024coreset,
  title={Coreset selection for object detection},
  author={Lee, Hojun and Kim, Suyoung and Lee, Junhoo and Yoo, Jaeyoung and Kwak, Nojun},
  booktitle={2024 IEEE/CVF Conference on Computer Vision and Pattern Recognition Workshops (CVPRW)},
  pages={7682--7691},
  year={2024},
  organization={IEEE}
}

@inproceedings{yang2024plug,
  title={Plug and play active learning for object detection},
  author={Yang, Chenhongyi and Huang, Lichao and Crowley, Elliot J},
  booktitle={2024 IEEE/CVF Conference on Computer Vision and Pattern Recognition (CVPR)},
  pages={17784--17793},
  year={2024},
  organization={IEEE}
}

@inproceedings{liu2023dream,
  title={Dream: Efficient dataset distillation by representative matching},
  author={Liu, Yanqing and Gu, Jianyang and Wang, Kai and Zhu, Zheng and Jiang, Wei and You, Yang},
  booktitle={2023 IEEE/CVF International Conference on Computer Vision (ICCV)},
  pages={17268--17278},
  year={2023},
  organization={IEEE}
}

@inproceedings{cui2019class,
  title={Class-balanced loss based on effective number of samples},
  author={Cui, Yin and Jia, Menglin and Lin, Tsung-Yi and Song, Yang and Belongie, Serge},
  booktitle={2019 IEEE/CVF conference on computer vision and pattern recognition (CVPR)},
  pages={9260--9269},
  year={2019},
  organization={IEEE}
}

@INPROCEEDINGS{tan2021eqlv2,
  author={Tan, Jingru and Lu, Xin and Zhang, Gang and Yin, Changqing and Li, Quanquan},
  booktitle={2021 IEEE/CVF Conference on Computer Vision and Pattern Recognition (CVPR)}, 
  title={Equalization Loss v2: A New Gradient Balance Approach for Long-tailed Object Detection}, 
  year={2021},
  volume={},
  number={},
  pages={1685-1694},
  doi={10.1109/CVPR46437.2021.00173}}

@InProceedings{zhao2025ltdd,
    author    = {Zhao, Zhenghao and Wang, Haoxuan and Shang, Yuzhang and Wang, Kai and Yan, Yan},
    title     = {Distilling Long-tailed Datasets},
    booktitle = {Proceedings of the IEEE/CVF Conference on Computer Vision and Pattern Recognition (CVPR)},
    month     = {June},
    year      = {2025},
    pages     = {30609-30618}
}

@inproceedings{cui2026rethinking,
  title={Rethinking long-tailed dataset distillation: A uni-level framework with unbiased recovery and relabeling},
  author={Cui, Xiao and Qin, Yulei and Li, Xinyue and Zhou, Wengang and Li, Hongsheng and Li, Houqiang},
  booktitle={Proceedings of the AAAI Conference on Artificial Intelligence},
  volume={40},
  number={5},
  pages={3443--3451},
  year={2026}
}

@INPROCEEDINGS{singh2018snip,
  author={Singh, Bharat and Davis, Larry S.},
  booktitle={2018 IEEE/CVF Conference on Computer Vision and Pattern Recognition}, 
  title={An Analysis of Scale Invariance in Object Detection - SNIP}, 
  year={2018},
  volume={},
  number={},
  pages={3578-3587},
  doi={10.1109/CVPR.2018.00377}}

@inproceedings{singh2018sniper,
 author = {Singh, Bharat and Najibi, Mahyar and Davis, Larry S},
 booktitle = {Advances in Neural Information Processing Systems},
 editor = {S. Bengio and H. Wallach and H. Larochelle and K. Grauman and N. Cesa-Bianchi and R. Garnett},
 pages = {},
 publisher = {Curran Associates, Inc.},
 title = {SNIPER: Efficient Multi-Scale Training},
 url = {https://proceedings.neurips.cc/paper_files/paper/2018/file/166cee72e93a992007a89b39eb29628b-Paper.pdf},
 volume = {31},
 year = {2018}
}

@inproceedings{ren2015faster,
 author = {Ren, Shaoqing and He, Kaiming and Girshick, Ross and Sun, Jian},
 booktitle = {Advances in Neural Information Processing Systems},
 editor = {C. Cortes and N. Lawrence and D. Lee and M. Sugiyama and R. Garnett},
 pages = {},
 publisher = {Curran Associates, Inc.},
 title = {Faster R-CNN: Towards Real-Time Object Detection with Region Proposal Networks},
 url = {https://proceedings.neurips.cc/paper_files/paper/2015/file/14bfa6bb14875e45bba028a21ed38046-Paper.pdf},
 volume = {28},
 year = {2015}
}

@INPROCEEDINGS{lin2017feature,
  author={Lin, Tsung-Yi and Dollár, Piotr and Girshick, Ross and He, Kaiming and Hariharan, Bharath and Belongie, Serge},
  booktitle={2017 IEEE Conference on Computer Vision and Pattern Recognition (CVPR)}, 
  title={Feature Pyramid Networks for Object Detection}, 
  year={2017},
  volume={},
  number={},
  pages={936-944},
  doi={10.1109/CVPR.2017.106}}

@inproceedings{carion2020end,
  title={End-to-end object detection with transformers},
  author={Carion, Nicolas and Massa, Francisco and Synnaeve, Gabriel and Usunier, Nicolas and Kirillov, Alexander and Zagoruyko, Sergey},
  booktitle={European conference on computer vision},
  pages={213--229},
  year={2020},
  organization={Springer}
}

@InProceedings{qi2025unidd,
    author    = {Qi, Ding and Li, Jian and Gao, Junyao and Dou, Shuguang and Tai, Ying and Hu, Jianlong and Zhao, Bo and Wang, Yabiao and Wang, Chengjie and Zhao, Cairong},
    title     = {Towards Universal Dataset Distillation via Task-Driven Diffusion},
    booktitle = {Proceedings of the IEEE/CVF Conference on Computer Vision and Pattern Recognition (CVPR)},
    month     = {June},
    year      = {2025},
    pages     = {10557-10566}
}

@article{tran2026rce,
  title={LLM-based Rarity-Aware Caption Expansion for Diffusion Dataset Condensation in Object Detection},
  author={Tran, Linh Tam and Bae, Sung-Ho},
  journal={Journal of Broadcast Engineering},
  volume={31},
  number={1},
  pages={162--168},
  year={2026}
}

@article{chen2026mgpo,
  title={MGPO: Manifold-Guided Diffusion Alignment for Task-Aware Dataset Distillation},
  author={Chen, Yunyi and Wang, Chenru and Ye, Xinyi and Zheng, Zexin and Zhang, Chi},
  journal={arXiv preprint arXiv:2610.05252},
  year={2026}
}

@inproceedings{
zhang2023dino,
title={{DINO}: {DETR} with Improved DeNoising Anchor Boxes for End-to-End Object Detection},
author={Hao Zhang and Feng Li and Shilong Liu and Lei Zhang and Hang Su and Jun Zhu and Lionel Ni and Heung-Yeung Shum},
booktitle={The Eleventh International Conference on Learning Representations },
year={2023},
url={https://openreview.net/forum?id=3mRwyG5one}
}

@ARTICLE{yu2024review,
  author={Yu, Ruonan and Liu, Songhua and Wang, Xinchao},
  journal={IEEE Transactions on Pattern Analysis and Machine Intelligence}, 
  title={Dataset Distillation: A Comprehensive Review}, 
  year={2024},
  volume={46},
  number={1},
  pages={150-170},
  doi={10.1109/TPAMI.2023.3323376}}

@InProceedings{kim2022idc,
  title = 	 {Dataset Condensation via Efficient Synthetic-Data Parameterization},
  author =       {Kim, Jang-Hyun and Kim, Jinuk and Oh, Seong Joon and Yun, Sangdoo and Song, Hwanjun and Jeong, Joonhyun and Ha, Jung-Woo and Song, Hyun Oh},
  booktitle = 	 {Proceedings of the 39th International Conference on Machine Learning},
  pages = 	 {11102--11118},
  year = 	 {2022},
  editor = 	 {Chaudhuri, Kamalika and Jegelka, Stefanie and Song, Le and Szepesvari, Csaba and Niu, Gang and Sabato, Sivan},
  volume = 	 {162},
  series = 	 {Proceedings of Machine Learning Research},
  month = 	 {17--23 Jul},
  publisher =    {PMLR},
  url = 	 {https://proceedings.mlr.press/v162/kim22c.html}
}

@InProceedings{du2023ftd,
    author    = {Du, Jiawei and Jiang, Yidi and Tan, Vincent Y. F. and Zhou, Joey Tianyi and Li, Haizhou},
    title     = {Minimizing the Accumulated Trajectory Error To Improve Dataset Distillation},
    booktitle = {Proceedings of the IEEE/CVF Conference on Computer Vision and Pattern Recognition (CVPR)},
    month     = {June},
    year      = {2023},
    pages     = {3749-3758}
}

@InProceedings{pooladzandi2022adacore,
  title = 	 {Adaptive Second Order Coresets for Data-efficient Machine Learning},
  author =       {Pooladzandi, Omead and Davini, David and Mirzasoleiman, Baharan},
  booktitle = 	 {Proceedings of the 39th International Conference on Machine Learning},
  pages = 	 {17848--17869},
  year = 	 {2022},
  editor = 	 {Chaudhuri, Kamalika and Jegelka, Stefanie and Song, Le and Szepesvari, Csaba and Niu, Gang and Sabato, Sivan},
  volume = 	 {162},
  series = 	 {Proceedings of Machine Learning Research},
  month = 	 {17--23 Jul},
  publisher =    {PMLR},
  url = 	 {https://proceedings.mlr.press/v162/pooladzandi22a.html}
}

@InProceedings{wu2022entropy,
    author    = {Wu, Jiaxi and Chen, Jiaxin and Huang, Di},
    title     = {Entropy-Based Active Learning for Object Detection With Progressive Diversity Constraint},
    booktitle = {Proceedings of the IEEE/CVF Conference on Computer Vision and Pattern Recognition (CVPR)},
    month     = {June},
    year      = {2022},
    pages     = {9397-9406}
}

@inproceedings{guo2024datm,
 author = {Guo, Ziyao and Wang, Kai and Cazenavette, George and LI, HUI and Zhang, Kaipeng and You, Yang},
 booktitle = {International Conference on Learning Representations},
 editor = {B. Kim and Y. Yue and S. Chaudhuri and K. Fragkiadaki and M. Khan and Y. Sun},
 pages = {56999--57020},
 title = {Towards Lossless Dataset Distillation via Difficulty-Aligned Trajectory Matching},
 url = {https://proceedings.iclr.cc/paper_files/paper/2024/file/fc740286fb4840cb56f99400817967da-Paper-Conference.pdf},
 volume = {2024},
 year = {2024}
}

@InProceedings{lee2024selmatch,
  title = 	 {{S}el{M}atch: Effectively Scaling Up Dataset Distillation via Selection-Based Initialization and Partial Updates by Trajectory Matching},
  author =       {Lee, Yongmin and Chung, Hye Won},
  booktitle = 	 {Proceedings of the 41st International Conference on Machine Learning},
  pages = 	 {26546--26567},
  year = 	 {2024},
  editor = 	 {Salakhutdinov, Ruslan and Kolter, Zico and Heller, Katherine and Weller, Adrian and Oliver, Nuria and Scarlett, Jonathan and Berkenkamp, Felix},
  volume = 	 {235},
  series = 	 {Proceedings of Machine Learning Research},
  month = 	 {21--27 Jul},
  publisher =    {PMLR},
  url = 	 {https://proceedings.mlr.press/v235/lee24g.html},
}

@inproceedings{lin2014microsoft,
  title={Microsoft coco: Common objects in context},
  author={Lin, Tsung-Yi and Maire, Michael and Belongie, Serge and Hays, James and Perona, Pietro and Ramanan, Deva and Doll{\'a}r, Piotr and Zitnick, C Lawrence},
  booktitle={European conference on computer vision},
  pages={740--755},
  year={2014},
  organization={Springer}
}

@article{everingham2010pascal,
  title={The pascal visual object classes (voc) challenge},
  author={Everingham, Mark and Van Gool, Luc and Williams, Christopher KI and Winn, John and Zisserman, Andrew},
  journal={International journal of computer vision},
  volume={88},
  number={2},
  pages={303--338},
  year={2010},
  publisher={Springer}
}

@inproceedings{rebuffi2017icarl,
  title={icarl: Incremental classifier and representation learning},
  author={Rebuffi, Sylvestre-Alvise and Kolesnikov, Alexander and Sperl, Georg and Lampert, Christoph H},
  booktitle={2017 IEEE conference on computer vision and pattern recognition (CVPR)},
  pages={5533--5542},
  year={2017},
  organization={IEEE}
}

@inproceedings{3023549.3023562,
author = {Chen, Yutian and Welling, Max and Smola, Alex},
title = {Super-samples from kernel herding},
booktitle = {Proceedings of the Twenty-Sixth Conference on Uncertainty in Artificial Intelligence},
year = {2010},
isbn = {9780974903965},
publisher = {AUAI Press},
address = {Arlington, Virginia, USA},
pages = {109--116},
numpages = {8},
location = {Catalina Island, CA},
series = {UAI'10}
}

@inproceedings{NEURIPS2022_631ad9ae,
 author = {Cao, Weihan and Zhang, Yifan and Gao, Jianfei and Cheng, Anda and Cheng, Ke and Cheng, Jian},
 booktitle = {Advances in Neural Information Processing Systems},
 doi = {10.52202/068431-1120},
 editor = {S. Koyejo and S. Mohamed and A. Agarwal and D. Belgrave and K. Cho and A. Oh},
 pages = {15394--15406},
 publisher = {Curran Associates, Inc.},
 title = {PKD: General Distillation Framework for Object Detectors via Pearson Correlation Coefficient},
 url = {https://proceedings.neurips.cc/paper_files/paper/2022/file/631ad9ae3174bf4d6c0f6fdca77335a4-Paper-Conference.pdf},
 volume = {35},
 year = {2022}
}
